\documentclass[10pt,journal,compsoc]{IEEEtran}
\newif\ifpeerreview

\peerreviewfalse

\usepackage[nocompress]{cite}
\usepackage{url}
\usepackage{amsmath,amssymb,graphicx}
\usepackage{hyperref}
\usepackage[switch]{lineno}
\usepackage{booktabs}
\usepackage{multirow}
\usepackage{mathtools}
\usepackage{subcaption}
\usepackage{cleveref}

\usepackage{pifont}
\usepackage[dvipsnames,table,xcdraw]{xcolor}
\usepackage{xspace}

\def\etal{\emph{et al.}}

\newcommand{\estcurrlf}{\hat{L}}
\newcommand{\lowrank}{\ensuremath{\mathbf{\mathcal{F}}}}

\newcommand{\xmark}{\scalebox{0.85}{\ding{53}}}%

\newcommand{\loss}{\mathcal L}
\newcommand{\warp}{\mathcal W}

\newcommand{\flownet}{\mathcal O}

\newcommand{\bfg}[1]{\textcolor{ForestGreen}{\textbf{#1}}}
\newcommand{\bfb}[1]{\textcolor{blue}{\textbf{#1}}}

\newcommand*\colourcheck[1]{%
  \expandafter\newcommand\csname #1check\endcsname{\textcolor{ForestGreen}{\ding{52}}}%
}

\newcommand*\colourmark[1]{%
  \expandafter\newcommand\csname #1mark\endcsname{\textcolor{#1}{\ding{55}}}%
}
\colourmark{red}
\colourcheck{blue}
\colourcheck{green}
\colourcheck{red}

\title{Stereo-Knowledge Distillation from \textit{dp}MV to Dual Pixels for Light Field Video Reconstruction}

\author{
Aryan Garg,
Raghav Mallampali, 
Akshat Joshi,
Shrisudhan Govindarajan, 
Kaushik Mitra \\   
Indian Institute of Technology, Madras
}

\begin{document}

\IEEEtitleabstractindextext{%

\begin{abstract}
Dual pixels contain disparity cues arising from the defocus blur. 
This disparity information is useful for many vision tasks ranging from autonomous driving to 3D creative realism. 
However, directly estimating disparity from dual pixels is less accurate. 
This work hypothesizes that distilling high-precision dark stereo knowledge, implicitly or explicitly, to efficient dual-pixel student networks enables faithful reconstructions. 
This dark knowledge distillation should also alleviate stereo-synchronization setup and calibration costs while dramatically increasing parameter and inference time efficiency. 
We collect the first and largest 3-view dual-pixel video dataset, dpMV, to validate our explicit dark knowledge distillation hypothesis. 
We show that these methods outperform purely monocular solutions, especially in challenging foreground-background separation regions using faithful guidance from dual pixels. 
Finally, we demonstrate an unconventional use case unlocked by dpMV and implicit dark knowledge distillation from an ensemble of teachers for Light Field (LF) video reconstruction. 
Our LF video reconstruction method is the fastest and most temporally consistent to date. 
It remains competitive in reconstruction fidelity while offering many other essential properties like high parameter efficiency, implicit disocclusion handling, zero-shot cross-dataset transfer, geometrically consistent inference on higher spatial-angular resolutions, and adaptive baseline control. 
All source code is available at the anonymous repository~\href{https://github.com/Aryan-Garg}{https://github.com/Aryan-Garg}.
\end{abstract}

}
\maketitle

\IEEEraisesectionheading{
  \section{Introduction}\label{sec:intro}
}
\IEEEPARstart{D}{ual} pixel (\textit{dp}) sensors have emerged as invaluable components in smartphone cameras, traditionally employed for auto-focusing~\cite{autofocus}. 
In contrast to monocular cameras, \textit{dp} sensors possess an inherent capability to capture disparity cues, thanks to a unique microlenslet array positioned over a single sensorboard. 
Recent advancements in \textit{dp}-based learning techniques~\cite{GargDualPixelsICCV2019_DPNet, Pan_2021_CVPR_DDDNet_DP_Disp}, along with other optimization methods \cite{punnappurath2020modeling, Xin_2021_ICCV_dual_pixel_cmu}, have facilitated the extraction of explicit disparity estimates directly from dual pixels, enabling numerous downstream vision tasks like autonomous driving and post-capture control~\cite{refocus_wadwa_18}.

However, directly estimating disparity from dual pixels presents challenges. 
While these sensors offer clear disparity cues in defocused regions (\cref{fig:scanline_dataset}), they struggle within the depth-of-field of the camera lens, thus reducing their effectiveness and necessitating an architectural understanding of depth-of-field and dynamic channel switching (RGB/DP).
Contrarily, this limitation does not exist for stereo-input methods and generally yields precise estimation, especially when trained on densely accurate large ground-truth datasets.
However, obtaining dense physical ground truth data from 3D laser scanning~\cite{kitti_dataset, eth_dataset_stereo} or structured light methods~\cite{middlebury} poses significant capturing and fidelity challenges. 
This motivated Garg \etal \cite{GargDualPixelsICCV2019_DPNet} to use multi-view stereo (MVS) computed disparities using \cite{colmap_COLMAP} for guidance. 
Still, MVS-computed disparities are inaccurate in textureless, occluded, and low-feature regions, leading to partial and sparse estimates,
inadvertently, introducing model bias and poor generalizability. 
A possible solution is to compute estimates from methods trained primarily on extensive \textit{synthetic} stereo datasets like SceneFlow~\cite{sceneflow_dataset}.
The synthetic ground truth is naturally dense, accurate, and more complete due to absolute scene control. 
Xu~\etal~\cite{xu2023unifying_unimatch} demonstrated a primarily synthetically trained method that achieved remarkable real-world domain adaptation.
\begin{figure}[t]
    \centering
    \includegraphics[width=0.98\columnwidth]{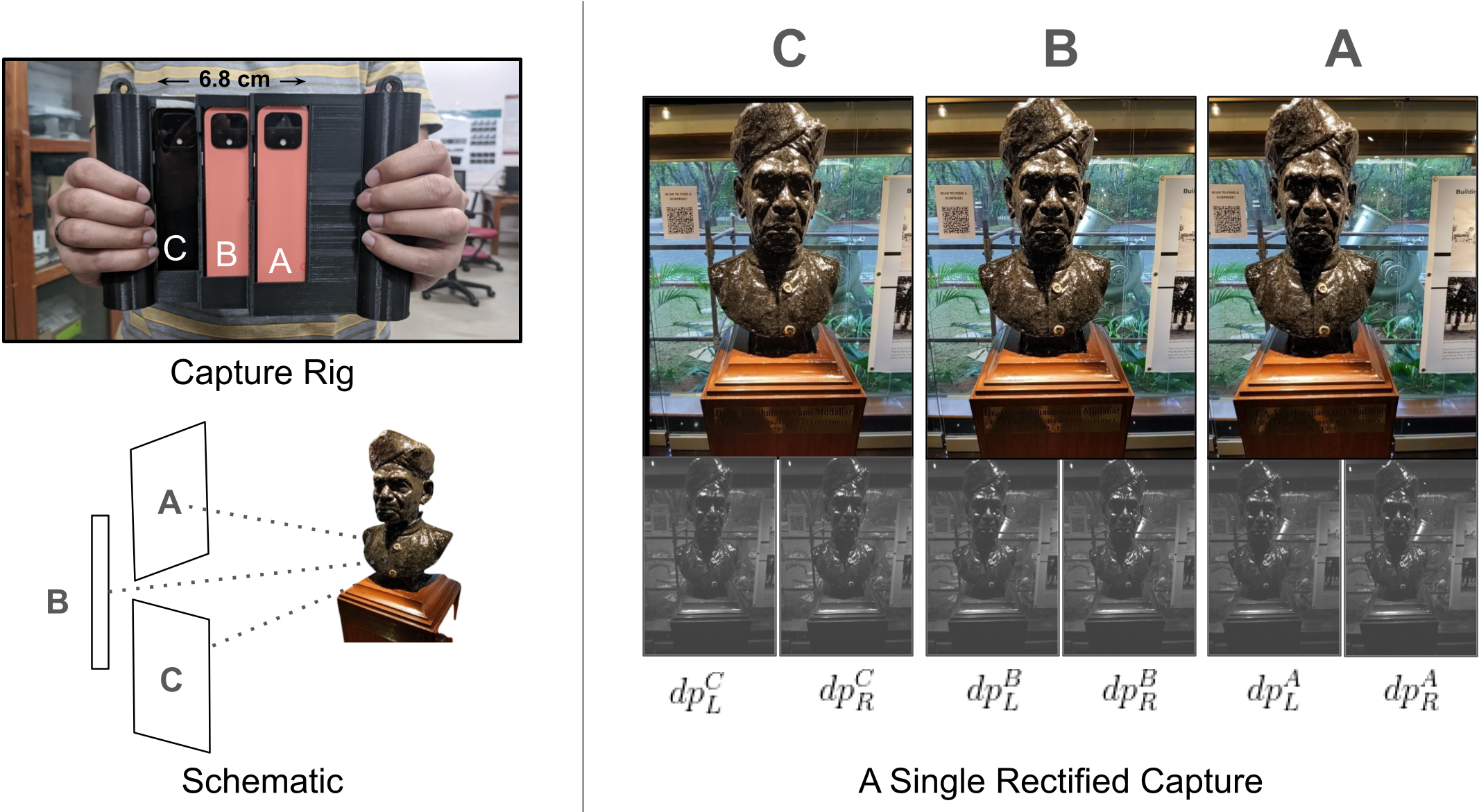}
    \caption{\textbf{\textit{dp}MV Dataset.} Capture rig and a single sample.}
    \label{fig:capture_single}
\end{figure}

While primary synthetic training and downstream real-world domain adaptation alleviate large prediction inaccuracies, integrating stereo cameras into smartphone devices still remains highly impractical. 
This is where dual pixels and our dark knowledge hypothesis bridge the gap, offering the best of both worlds: high synthetic pre-training accuracy without any hardware overhead.
We propose using \textit{dark knowledge} distillation~\cite{kd_first_01, kd_survey_02} from a synthetically pre-trained stereo teacher to a dual pixel input network.

\begin{figure*}[t]
    \centering
    \includegraphics[width=0.99\textwidth]{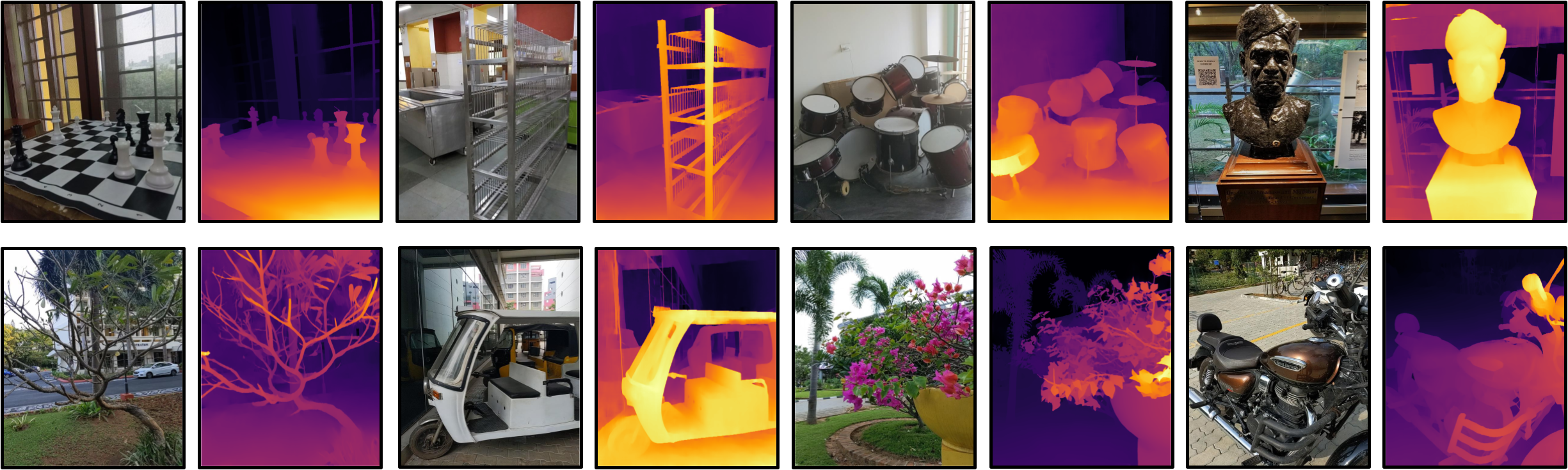}
    \caption{\textbf{Scenes of \textit{dp}MV} with computed stereo-disparity. Top row: indoor scenes. Bottom row: outdoor scenes}
    \label{fig:dataset_samples}
\end{figure*}

\begin{table*}[ht]
    \centering
    \caption{\textbf{Multi-View and Dual Pixel Datasets.} Images are counted for 1-viewpoint only. ETH3D-LRMV corresponds to ETH3D's Low-Resolution Multi-View Split.}
    \begin{tabular}{lccccccccc}
        \toprule
         Dataset & Views & Resolution & has~\textit{dp} & Scenes & Images & has Video & \textit{fps} & Setting & GT-Disparity \\
        \midrule
        
        Middlebury-14~\cite{middlebury} & 2 & 640$\times$480 & \xmark & 33 & 33 & \xmark & \xmark & Indoor  & St. Light \\    
        KITTI-15~\cite{kitti_dataset} & 2 & 1242$\times$375 & \xmark & 400 & 400 &  \xmark & \xmark & Street  & 3D-Laser \\     
        ETH3D-LRMV~\cite{eth_dataset_stereo} & 4 & 752$\times$480 & \xmark & 10 & 10,008 & \checkmark & 13 & Varied  & 3D-Laser \\ 
        SceneFlow~\cite{sceneflow_dataset} & 2 & 960$\times$540 & \xmark & 2256 & 34,801 & \xmark & \xmark & Synth. &  Synthetic \\
        
        Garg~\etal~\cite{GargDualPixelsICCV2019_DPNet} & \textbf{5} & 1512$\times$2016 & \checkmark & \textbf{3,573} &  17,865 & \xmark  & \xmark & Varied & Computed \\ 
        \midrule
         \textbf{\textit{dp}MV(Ours)} & 3 & \textbf{1512$\times$2016} & \checkmark & 145 & \textbf{58,295} & \checkmark & \textbf{13} & Varied & Computed  \\
        \bottomrule
    \end{tabular}
    \label{tab:dataset_compare}
\end{table*}
To validate this hypothesis, we collected the dual-pixel multiview videos (\textit{dp}MV) dataset – the first and largest 3-view dual-pixel video dataset (\cref{fig:capture_single}, \cref{fig:dataset_samples}) captured in 145 varied indoor-outdoor scenarios that allow dark stereo knowledge distillation for dual pixel input networks. 
See~\cref{tab:dataset_compare} for dataset features and comparison. 
We demonstrate in~\cref{fig:dp_disp_comparison} that methods incorporating dark knowledge outperform purely dual-pixel solutions~\cite{Pan_2021_CVPR_DDDNet_DP_Disp} as well as massively trained monocular solutions~\cite{depth_anything, midas, Ranftl2020}, particularly in challenging foreground-background separation where guidance from dual pixels proves indispensable.

Furthermore, we explore an unconventional application unlocked by dpMV and the dark knowledge distillation hypothesis: light field video reconstruction from a single view.
Light fields~\cite{MarcLevoy_LFRendering}, represent the amount of light traveling in every spatial-angular direction, allowing for a comprehensive description of a scene's visual properties essential for applications in various fields, including post-capture control~\cite{refocus_wadwa_18} and virtual reality~\cite{khan2021edgeaware, khan2020vclfd}.
Traditional and multi-camera array methods of capturing light fields are expensive, impractical, slow, and require expertise to operate~\cite{ng2005light, lowCostMultiCameraArray, systemForAcquiringLF_google, broxton2020immersive_google}. 
To address these challenges, researchers have turned to machine learning for light field (LF) reconstruction from sparse views~\cite{eccv_22_monoLFVR, Li_VMPI, Srinivasan_2017_ICCV_4D_RGBD, 5d_monoVideo_recons_2019, ivan2019synthesizing, LFGAN_2023, Selfvi, LearningViewSynthesis_SIG16, Niklaus_Ken_Burns} (Summarized in \cref{tab:related}).
These methods typically rely on convolutional neural network (CNN) backbones that miss dense and globally dependent visual information.
Furthermore, while Mono~\cite{eccv_22_monoLFVR} is a top-performing single-view reconstruction method aimed at smartphones, it relies heavily on external depth-estimation networks for context, requires ground truth LFs for refinement, and produces physically inaccurate reconstructions at boundary conditions (\cref{fig:quan_eval_resAgnos_and_dp_benefits}).

We propose a novel cross-modal self-supervised~\cite{rebuttal_01, rebuttal_02} approach for LF video reconstruction using \textit{just} dual pixels and our dark knowledge hypothesis that alleviates Mono's limitations. 
We use vision transformers~\cite{dosovitskiy2020vit} for the first time in LF reconstruction that captures dense features with simultaneous global and local context enabling many zero-shot properties of our method.
Leveraging our \textit{implicit} dark knowledge distillation framework through an ensemble of disparity~\cite{xu2023unifying_unimatch} and optical flow~\cite{raft_RAFT} teachers, our proposed method achieves the fastest inference speed and highest temporal consistency. 
Our method offers many other crucial advantages for practical reconstruction: high parameter efficiency, implicit disocclusion handling, zero-shot cross-dataset transfer, adaptive baseline control, and robust geometric consistency. 
In summary, we make the following contributions:
\begin{enumerate}
    \item \textbf{dpMV: Dual Pixel Multi-Views Video Dataset.} We collect the \textit{first} and \textit{largest} dual-pixel 3-view video dataset that also enables our proposed dark knowledge hypothesis.
    It contains 58,295 frames per viewpoint, spanning 145 varied indoor and outdoor challenging scenes.
    The dataset also has classes like petals, chairs, tables, etc. enabling classification, automatic video captioning, and other multi-modal learning tasks.
    
    \item \textbf{Dark Knowledge Distillation Hypothesis and \textit{dp}MV Benchmark.} We redefine dark knowledge~\cite{kd_first_02} in~\cref{sec:SDKM3} and then propose the hypothesis.
    We validate it by providing a disparity-estimation benchmark on \textit{dp}MV.
    \item \textbf{Novel Light Field Video Reconstruction (LFVR) Method.} We introduce the \textit{first method} that reconstructs light field videos from dual pixels. 
    
    \item \textbf{Vision Transformers for LFVR.} We demonstrate the \textit{first use case} of vision transformers as the primary backbone for LFVR.
    \item \textbf{Fastest and most temporally consistent LFVR.} Our method is the fastest at $159\, ms$ per reconstruction and uses a meagre $38.18M$ parameters. Our method also achieves the highest temporal consistency and competitive fidelity scores.
\end{enumerate}

\section{Related Work}
\label{sec:related_work}

\subsection{Disparity Estimation}
The monocular baselines used for qualitative comparison in \cref{fig:dp_disp_comparison} are DepthAnything \cite{depth_anything}, MiDaS \cite{midas} and DPT \cite{Ranftl2020}.
\textit{dp}-based baselines include DDDNet \cite{Pan_2021_CVPR_DDDNet_DP_Disp} and DPDD \cite{punnappurath2020modeling}. See supplementary for extended related work.


\subsection{Knowledge Distillation}
Knowledge distillation (KD) was first introduced in model compression~\cite{kd_first_01} and then more explicitly by Hinton~\etal~\cite{kd_first_02}.
Inspired by the demonstration that shallow feed-forward networks can learn complex functions~\cite{do_nn_need_to_be_deep}, extensive research has been conducted on knowledge distillation~\cite{kd_survey_01, kd_survey_02}. 
This includes response-based~\cite{response_01, response_02, response_03, response_04}, feature-based~\cite{feature_01, feature_02}, and contrastive techniques~\cite{contrastive_01}.
In this paper, we focus on implicit and explicit response-based distillation techniques.
\subsection{Vision Transformers.}
Vision Transformers (ViT~\cite{dosovitskiy2020vit}) have shown state-of-the-art performance across domains however remain compute-intensive due to no inductive biases of locality. 
Data efficient transformers like DeiT~\cite{data_eff_transformers_deit} explore distillation from other large ViTs but still lack the locality cues. 
So, we focus on the class of convolution-infused vision transformers~\cite{conVit_01, conVit_02, conVit_03, conVit_04, wu2021cvt} that explicitly enforce the locality bias~\cite{early_convolutions, token_to_vit} for better spatial-feature modeling and simultaneously reduction of the quadratic complexity of attention. 
Our Scene Decomposer ViT is inspired by CvT~\cite{wu2021cvt} and attention with residual connections~\cite{conVit_03}. 

\subsection{Light Field Capturing and Synthesis}
Raytrix and Lytro Illum~\cite{ng2005light} are traditional capturing cameras that are expensive, slow, not agnostic to large motions, and limited to expert users.
Renewed interest in LF capture, processing, rendering \cite{lowCostMultiCameraArray, systemForAcquiringLF_google, broxton2020immersive_google} has been observed.
However, multi-camera array setups are difficult to scale and deploy, making the technology inaccessible. 
Machine learning has been employed for reconstructing LFs from sparse (1, 2 and 4) views~\cite{eccv_22_monoLFVR, Li_VMPI, Srinivasan_2017_ICCV_4D_RGBD, 5d_monoVideo_recons_2019, ivan2019synthesizing, LFGAN_2023, vldi_Bak_2023, Selfvi, bino-LF, LearningViewSynthesis_SIG16, Niklaus_Ken_Burns}. See~\cref{tab:related} for a concise summary.
Current reproducible state-of-the-art (fidelity) methods are self-supervised approaches: SeLFVi~\cite{Selfvi} and Mono~\cite{eccv_22_monoLFVR} 
\begin{table}[h]
    \centering
    \caption{\textbf{Summary of sparse-view light field reconstruction related work}. End-to-end ($E2E$) means no external network computed input requirements. Input ($In-Views$) denotes the number of input views required and $SS$ stands for self-supervised.}
    \begin{tabular}{l|ccc}
    \toprule
    Method & SS  & In-Views & E2E  \\
    \midrule
    RF - View synthesis \cite{mildenhall2020nerf, liu2021neural, zhang2020nerf} & \checkmark & $\ge$ 4 & \xmark \\
 
    RF - View synthesis \cite{Xu_2022_SinNeRF, yu2021pixelnerf} & \checkmark & 1 & \xmark \\

     LF synthesis \cite{LearningViewSynthesis_SIG16,wu2017light,wang2018end, LocalLF_fusion_2020, flynn2019deepview} & ~~\xmark~~ & $\ge$ 4 & ~~\checkmark~~  \\
  
    X-fields~\cite{Bemana2020xfields} & \checkmark & 4 & \checkmark \\   

    SelFVi~\cite{Selfvi} \& Bino-LF \cite{bino-LF} & \checkmark & 2 & \checkmark \\ 
 
    MPI \& LDI based \cite{Li_VMPI, vldi_Bak_2023, single_view_mpi} & \xmark & 1 & \xmark \\   

    5D-LF \cite{5d_monoVideo_recons_2019}, LFGAN~\cite{LFGAN_2023} \& SMPI~\cite{SMPI} & \xmark & 1 & \checkmark \\
       
    Mono~\cite{eccv_22_monoLFVR} & \checkmark & 1 & \xmark \\  
    \midrule
    \textbf{Ours} & \checkmark & 1$+dp$ & \checkmark \\    
    \bottomrule
    \end{tabular}    
    \label{tab:related}
\end{table}
\section{\textit{dp}MV: Dual Pixel Multiview Videos Dataset}
\label{sec:dataset}
In this section, we introduce \textit{dp}MV, the first and largest dual-pixel video dataset for stereo and \textit{dp}-vision tasks with our dark knowledge hypothesis. It also has classes per video enabling classification, automatic video captioning, and novel multi-view + dual pixel multi-modal learning tasks.

\subsection{Hardware and Capturing.}
\label{subsec:capture-set-up}
The dataset is captured using a custom-manufactured rig with three slots ($A$, $B$, $C$), each fitted with a Google Pixel 4 smartphone. 
The central phone initiates contact with the SNTP server by sending a timestamp to capture synchronized videos. 
Due to this synchronization, videos are restricted to a maximum of 13 frames per second ($fps$). 
The spatial resolution of capturing is 1512$\times$2016.

\subsection{Defocus cues in dual pixels}
\begin{figure}[ht]
    \centering
    \includegraphics[width=0.99\columnwidth]{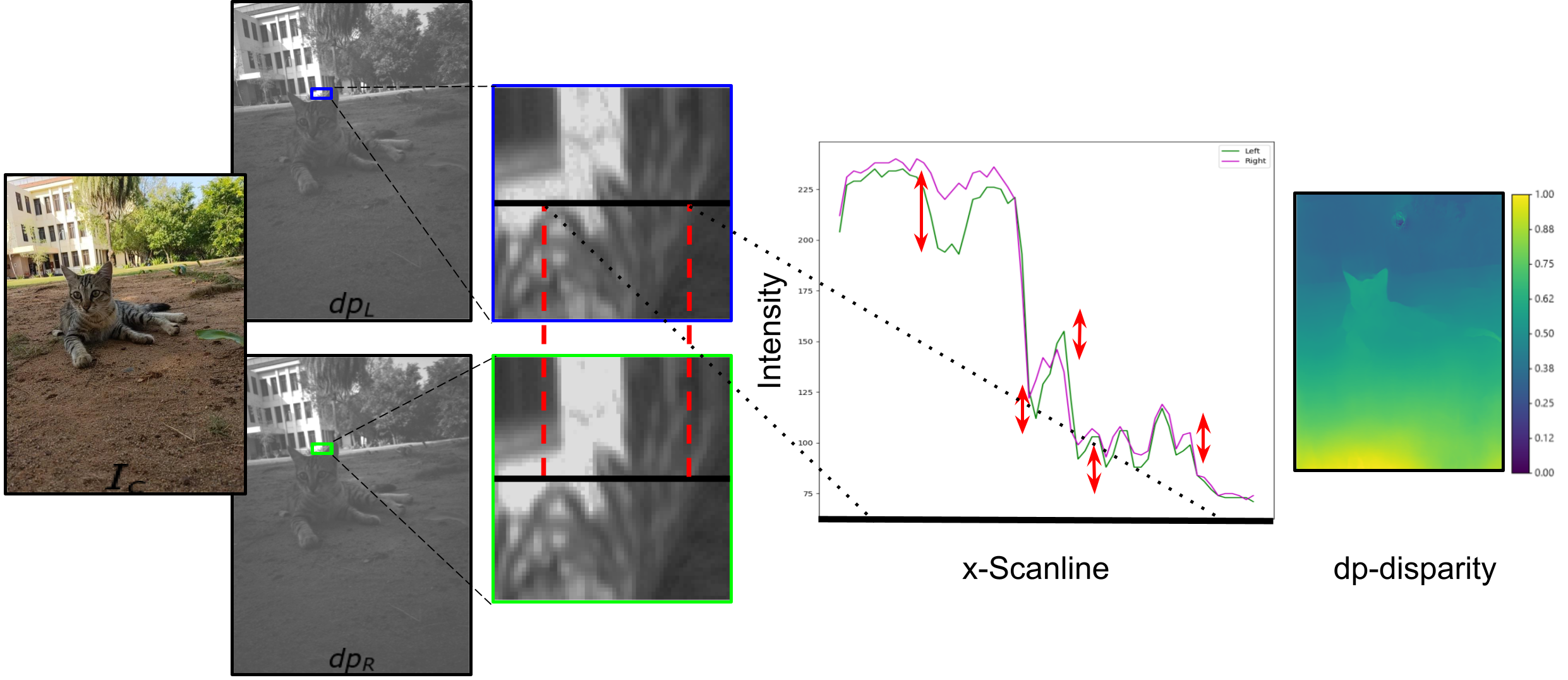}
    \caption{Scanline histogram analysis of dual pixels for defocus cues in foreground-background separation regions. Shown dp-disparity is from our best-fidelity method.}
    \label{fig:scanline_dataset}
\end{figure}

In~\cref{fig:scanline_dataset}, we analyze foreground-background separation regions for defocus or disparity cues. 
The intensity differences are most prominent in these regions which are crucial for sharp disparity estimates.
Despite an \textit{almost} all-in-focus capture, dual pixels still contain valuable information. 

\begin{figure*}[ht]
    \centering
    \includegraphics[width=0.99\textwidth]{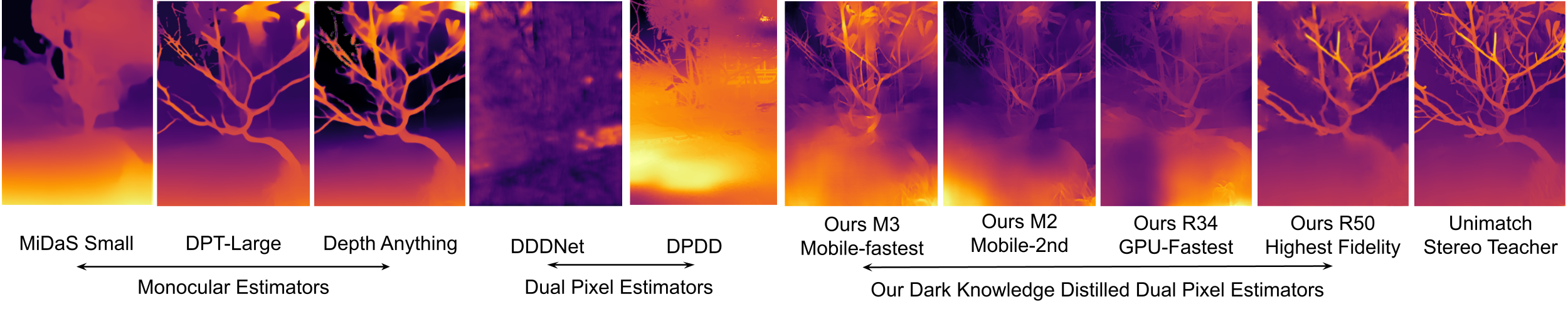}
    \caption{\textbf{Hypothesis Validation.}
    Dark knowledge distilled \textit{dp}-estimators outperform monocular and purely \textit{dp} methods. 
    Almost all-in-focus outdoor~\textit{dp}MV scene is chosen.}
    \label{fig:dp_disp_comparison}
\end{figure*}

\begin{table*}[ht]
    \caption{\textbf{Quantitative Comparison and \textit{dp}MV Benchmark} with our dark \textit{dp}-disparity estimators. The top section shows monocular estimators~\cite{midas, Ranftl2020, depth_anything}, the second division shows \textit{dp}-baselines~\cite{Pan_2021_CVPR_DDDNet_DP_Disp, punnappurath2020modeling} and the rest demonstrates our methods.
    Even though MiDaS~\cite{midas} is significantly smaller and faster than DPT-Large-384~\cite{Ranftl2020}, it is prone to a massive fidelity drop of $\sim 76\%$.
    Contrarily, our methods provide the highest fidelity at real-time inference speeds and lowest parameters.
    All wall clock times ($W_{GPU}$ and $W_{CPU}$), per estimation, are in mili-seconds measured on a single NVIDIA T4 GPU.
    Parameters ($Params$) are in millions. 
    }
    \centering
    \begin{tabular}{lccccccc}
    \toprule
        Method & AI (1) $\downarrow $ & AI (2) $\downarrow $ & Params. ($M$)  & $W_{CPU}$ & $W_{GPU}$  \\
        \midrule \midrule
        Depth-Anything~\cite{depth_anything} & 0.212 & 0.261 & 335.3 & $>$20000 & 72.38 \\     
        DPT-Large~\cite{Ranftl2020} & 0.196 & 0.231 & 344.05 & 11770.71 & 17.63 \\  
        MiDaS~\cite{midas} & 0.237 & 0.304 & 21.32 & 427.31 & 13.82 \\     
        \midrule  
        DDDNet~\cite{Pan_2021_CVPR_DDDNet_DP_Disp} & 0.298 & 0.389  &  10.964  & 4223.05 & 73.34 \\    
        DPDD~\cite{punnappurath2020modeling} & 0.221 & 0.341 & - & $>$20000 & - \\  
        \midrule
        \textbf{Ours-Mv2} & 0.148 & 0.191 & \textbf{6.824} & 193.21 & 15.04 \\
        \textbf{Ours-Mv3-CPU}  & 0.182 & 0.215 & 6.916  & \textbf{173.96} & 17.71 &  \\ 
        \midrule
        \textbf{Ours-R34-{GPU}} & 0.129 & 0.169 &  26.08 & 538.26 & \textbf{11.20} \\
        \textbf{Ours-R50-Best} & \textbf{0.129} & \textbf{0.165}  & 48.98 & 1508.89 & 21.67 \\ 
    \bottomrule
    \end{tabular}
    \label{tab:benchmark_dpmv}
\end{table*}

\subsection{Statistics, Scenes, and Comparison}
\label{sec:scenes-dataset-stats}
\textit{dp}MV encompasses 145 diverse indoor and outdoor scenes, resulting in 58,295 RGB+\textit{dp} frames per device. 
Each frame has 5 channels, totaling $58295\times 3\times 5$. 
Disparity maps are computed for each $B-A$ pair and serve as \textit{dark knowledge} for both, disparity estimation and light field reconstruction. 
\textit{dp}MV scenes encompass diverse indoor and outdoor environments, such as stores, buildings, museums, and forested areas, providing rich learning features (\cref{fig:dataset_samples}).
All classification targets are provided.
They exhibit varying illumination levels, disocclusion scenarios, and surfaces ranging from metallic and reflective to transmissive. 
Visible light sources add to the dataset's complexity.
Videos with light sources, non-Lambertian surfaces, and reflective or transmissive surfaces are further categorized.

\subsection{Dark Stereo Knowledge Distillation Hypothesis}
\label{sec:SDKM3}

\textbf{Motivation.} 
Dual pixel sensors excel in defocused regions but struggle in focused areas.
This limits their effectiveness, especially with inaccurate disparity supervision.
Synthetic stereo-disparity estimators are densely accurate but require careful hardware synchronization and calibration in real-world operating conditions. 
Both limitations, hardware synchronization, and inaccurate estimations, can be simultaneously overcome by distilling \textit{dark knowledge} from a stereo network to a \textit{dp}-network.

\textbf{Redefining Dark Knowledge}
We define \textit{dark knowledge}, as first described in~\cite{kd_first_02}, to encompass information that a student network can not discover autonomously even if no dissimilarity in networks' representation power existed.
In our context, this includes the stereo-disparity computed from an \textit{additional} view in \textit{dp}MV. 
This constitutes undiscoverable information for the student network as the student accesses only one view with \textit{dp} channels.

\textbf{Proposed Hypothesis.}
Response-based distillation of affine-invariant dark knowledge from a synthetically trained stereo-teacher to a single view \textit{dp}-student allows estimating dense disparity without the requirement of an additional camera or stereo-view.
Recent findings indicate that response-based students learn correlations in the data and shape their latent space that approximates the teacher's, rather than naively mimicking outputs~\cite{what_knowledge_ojha}. 
Additionally, students can surpass their teachers in task accuracy~\cite{student_better_than_teacher}.
So, theoretically, a student can refine estimates along challenging foreground-background edges by leveraging defocus cues (\cref{fig:scanline_dataset}) while learning to dynamically switch between RGB and \textit{dp}-channels when disparity cues are absent or ambiguous for in-focus regions.
Intuitively, the student is a camera-focus-aware estimator. 
See~\cref{fig:dp_disp_comparison} for hypothesis validation.

\begin{figure}[ht]
    \centering
    \includegraphics[width=0.99\columnwidth]{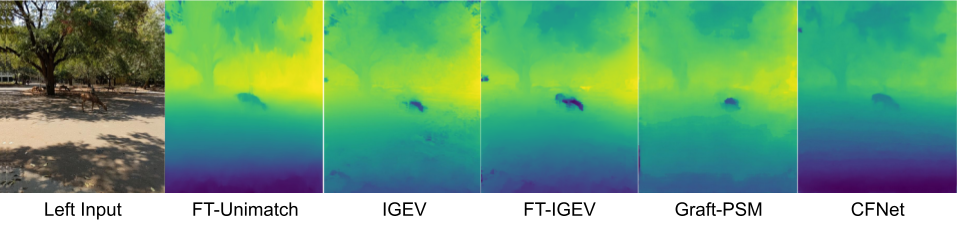}
    \caption{\textbf{Dark Knowledge Stereo Disparity Teacher Selection.} $FT$ stands for hyper-parameter fine-tuned. 
    The stereo teacher selected here (Unimatch~\cite{xu2023unifying_unimatch}) is also used as the geometry teacher for our light field video reconstruction method (\ref{sec:lfvr})}
    \label{fig:teacher_selection}
\end{figure}

\textbf{Validation of Dark Knowledge Distillation Hypothesis} 
We select the SceneFlow~\cite{sceneflow_dataset} pre-trained teacher Unimatch~\cite{xu2023unifying_unimatch} for training students after a qualitative comparison shown in~\cref{fig:teacher_selection}.
Note that ACVNet~\cite{xu2022attention_ACVNet} and STTR~\cite{STTR} were also tried on the scenes of \textit{dp}MV but performed poorer than the shown methods.

Our tiny-student-nets are $4$ off-the-shelf encoders~\cite{resnet, mv2, mv3} within an adapted 5-layer-depth U-Net++~\cite{unet_pp} that yield high quality disparity~\cref{fig:dp_disp_comparison}.
These tiny networks (smallest at $6.8M$ parameters) allow fast processing times of 173 ms on a CPU and 11.2 ms ($\sim90 fps$) on an NVIDIA T4 GPU.
Note that T4 GPU is slightly faster than GTX 1050, comparable to Snapdragon's Adreno 750 for smartphones.

\begin{figure}[ht]
    \centering
    \includegraphics[width=0.8\columnwidth]{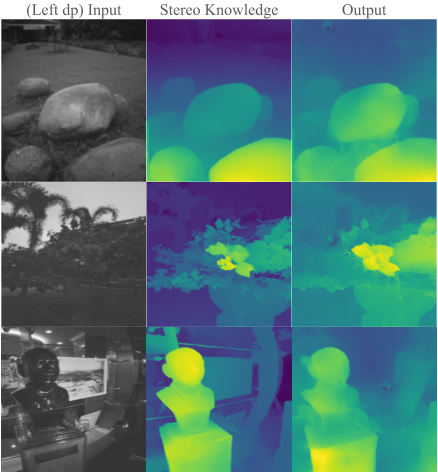}
    \caption{\textbf{\textit{dp}-Disparity Estimation Qualitative Results} from our best fidelity network: Dark-R50-Best (\cref{tab:benchmark_dpmv}).}
    \label{fig:dp_disp_more}
\end{figure}
We also \textit{quantitatively} benchmark~\textit{dp}MV against the same baselines as in~\cref{fig:dp_disp_comparison} (\cite{midas, Ranftl2020, depth_anything, Pan_2021_CVPR_DDDNet_DP_Disp, punnappurath2020modeling}) and our proposed tiny-student-nets, using the teacher's stereo-disparity as the reference (affine-invariance is ensured) in~\cref{tab:benchmark_dpmv}. 
Our best-fidelity qualitative estimations are shown in~\cref{fig:dp_disp_more}.

\textbf{Evaluation Metrics}
\label{sec:affine_inv_metrics}       
We use affine invariant MAE ($AI(1)$) and affine invariant RMSE ($AI(2)$) for evaluation, similar to~\cite{punnappurath2020modeling} in~\cref{tab:benchmark_dpmv}.
Parameters and inference time efficiencies are essential factors for real-world applications and edge-device deployments, necessitating consideration within our study, as shown in~\cref{tab:benchmark_dpmv}.

\section{Light Field Video Reconstruction from Dual Pixels}
\label{sec:lfvr}
Our \textit{self-supervised} light field (LF) video reconstruction methodology, architecture, and losses are described in this section.
\begin{figure*}[ht]
    \centering
    \includegraphics[width=\textwidth]{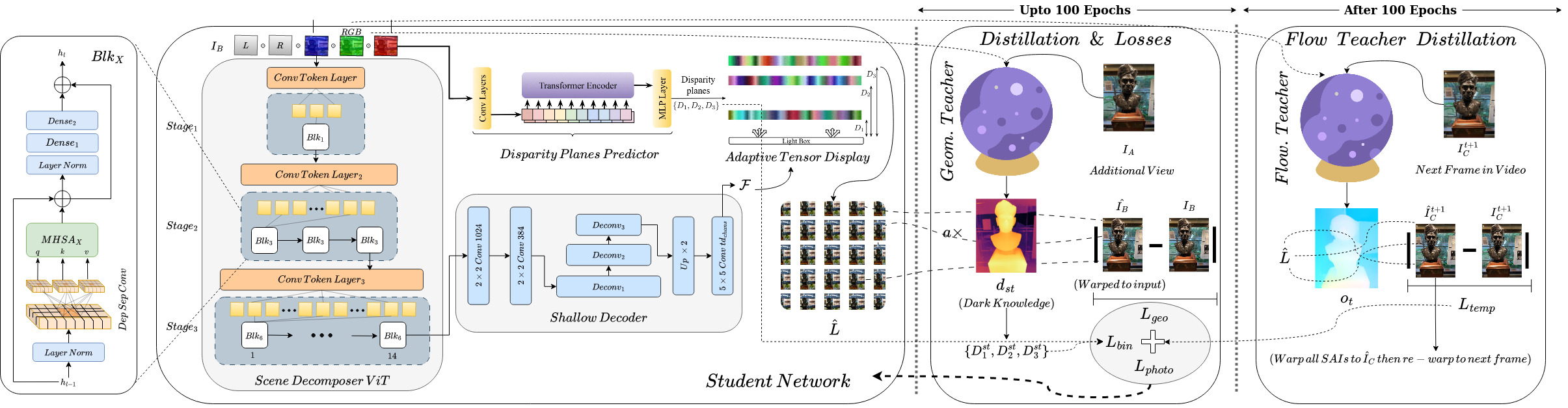}
    \caption{\textbf{Model Architecture.} 
    \textit{Left.} shows the architecture components: scene decomposer ViT and shallow decoder for a low-rank LF representation ($\lowrank$), the disparity planes predictor ViT, and the parameter-free adaptive tensor display layer.
    \textit{Right.} shows the stereo teacher's dark knowledge for warping sub-aperture views back to the input view ($I_B$) and estimating true scene disparity planes for the geometric and chamfer distance losses. 
    The dark flow teacher calculates the temporal distillation loss ($L_{temp}$), described in~\cref{subsec:losses}.
    }
    \label{fig:main_pipeline}
\end{figure*}

Given a single view ($\{I_B,dp^{B}_{L}\,dp^{B}_{R}\}$) where $B$ denotes the central camera in \textit{dp}MV), our goal is to synthesize novel views in $u \times v$ angular directions, or reconstruct a light field $\estcurrlf \in \mathbb{R}^{u \times v \times x \times y}$ where $x \times y$ represents the spatial dimensions.
Our method consists of two parallel ViTs responsible for decomposing scene features and extracting accurate disparity planes from the same input.
Using the disparity planes and the \textit{decoded} scene decomposition ($\lowrank$), an angular-grid structure is imposed on $\lowrank$ for tensor display~\cite{TD_layer}, similar to~\cite{Selfvi, eccv_22_monoLFVR}.
This returns a $u \times v$ structured light field ($\hat{L}$) from the single input view.  
We use an \textit{implicit} dark knowledge distillation framework here with an ensemble of two teachers: Geometry-teacher (Unimatch~\cite{xu2023unifying_unimatch}) and Flow-teacher (RAFT~\cite{raft_RAFT}). 
The term \textit{implicit} is used because our proposed student does not explicitly estimate either disparity or flow. 
Instead, it is trained to generate LFs that closely match the teacher's computed disparity baseline ($d_{st}$) and produce consistent sub-aperture images (SAIs) that warp to the teacher's dark knowledge input ($I^B_{t+1}$) accurately using the flow-teacher's dark knowledge (optical flow estimate: $o_t$). 
The student is initially trained solely by the geometry teacher for a fixed number of iterations before the flow-teacher ensemble distillation begins.

It is imperative to note the clear distinction of using the contextual information provided by an external network on the feedback side versus for input as done in most previous single-view LF reconstruction works~\cite{eccv_22_monoLFVR, 5d_monoVideo_recons_2019, LFGAN_2023, single_view_mpi, Li_VMPI}. 
This vital difference alleviates critical issues like external input side dependencies, and massive inference latencies due to an additional network (\cref{tab:temp_params_speed}). 
Fidelity ceilings and geometric consistency can be attributed to these external networks, as shown in \cref{fig:quan_eval_resAgnos_and_dp_benefits}. 
Our student architecture and geometry-teacher training stage is shown in~\cref{fig:main_pipeline}.

\subsection{Scene Decomposer Vision Transformer}
In subsequent text, we formally describe the formation of a low-rank light field representation ($\lowrank$) from the dual pixel input ({$I_B, dp^B_L, dp^B_R$}) using the scene decomposition ViT and shallow deconvolutional decoder.

\textbf{Tokenization.} 
The 5-channel input ($x_0 := \{I_B, dp_L, dp_R\} \in \mathbb{R}^{H \times W \times 5}$), undergoes processing through a convolutional layer $f^1_{tok}(\cdot)$. This layer employs a large $k \times k$ kernel size, to capture low-level features and reduce the quadratic complexity burden of self-attention. 
The resulting spatial map $f^1_{tok}(x_0) \in \mathbb{R}^{H_1\times W_1 \times C_{tok_1}}$ is then flattened into $H_1W_1 \times C_{tok_1}$ and forwarded to the first vision transformer stage without positional embeddings similar to~\cite{wu2021cvt}.

In contrast to traditional vision transformers that operate on fixed-size image patches in the pixel space, these convolutional layers eliminate the early weak inductive bias of locality in vision transformers \cite{early_convolutions, bridging_the_gap_bw_ViT_and_CNNs, conVit_04}, thereby dispensing with the need for positional embeddings. 
Moreover, convolutional tokenization helps avoid discontinuous patch embeddings observed in vanilla attention-based networks~\cite{token_to_vit, conVit_04}, which could be detrimental for dense prediction tasks such as LFVR.
To maintain control over spatial and feature dimensions, we incorporate this convolutional tokenization process before each ViT stage on the 2D-reshaped token map ($x_i \in \mathbb{R}^{H_i \times W_i \times C_{tok_{i}}}$). 
By increasing the feature channels ($C_{tok_i} > C_{tok_{i-1}}$) and reducing the spatial dimension ($H_i < H_{i-1}$) akin to CNNs, we aim to capture increasingly complex semantic features efficiently.

\textbf{Self-Attention Blocks.} 
Each ViT stage receives the convolved-flattened tokens ($x_i \in \mathbb{R}^{H_iW_i \times C_{tok_i}}$) and passes them onto sequential blocks ($Blk_X$ in~\cref{fig:main_pipeline}). 
Each block follows a similar architecture where $x_i$ is first passed through a depth-wise separable convolution layer from~\cite{depsepconv_xception} (Dep Sep Conv in~\cref{fig:main_pipeline}), instead of the conventional feed-forward query ($Q$), key ($K$), and value ($V$) MLPs.
These convolutions further model local context and reduce the burden of quadratic complexity of self-attention computation described subsequently. 
Each self-attention head performs the standard softmax scaled dot product~\cite{dosovitskiy2020vit} between a query ($x^q_i$), a key ($x^k_i$), and a value tensor ($x^v_i$). 
We use squeezed self-attention, halving the dimension of K-V pair.
This is based on the sparse self-attention rank observation by Wang~\etal~\cite{Sinong_lowrank}.
After attention modeling, residual connections to propagate it follow, similar to~\cite{conVit_03}. Conventional layer normalization and two dense layers are added to avoid vanishing gradients, distribution skewing, and rank collapse~\cite{Dong_rank_collapse}.

\textbf{No CLS token.} 
After repeating the core tokenization and MHSA blocks across 3 ViT stages and 1, 3 and 14 blocks in each, we pass the output latent 2D-reshaped token map directly to our shallow deconvolutional decoder instead of learning a representative $CLS$ token as shown in~\cref{fig:main_pipeline}.

\textbf{Low-Rank LF Representation Decoding}
We use the latent representation obtained from the Scene Decomposition ViT as input for the shallow deconvolutional decoder which returns a low-rank LF representation $\lowrank \in \mathbb{R^{L\times M}}$. 
Each channel in $\lowrank$ can be intuitively thought of as a scene depth plane, much like the multi-plane image paradigm~\cite{single_view_mpi, Li_VMPI}. 
Formally, $\lowrank = [\mathbf{f}_{-L/2}, \ldots, \mathbf{f}_{0},  \ldots, \mathbf{f}_{L/2} ]$,
where $\mathbf{f}_k = [f_k^1, f_k^2, \ldots, f_k^M]^T$, $f_k^m\in [0,1]^{h \times w \times 3}$ consists of $L\times M$ RGB channels. 
$L$ and $M$ are the number of layers and the rank for the adaptive tensor display~\cite{TD_layer}. 
This layer also requires scene disparity planes.


\subsection{DPP: Disparity Planes Predictor.}
\label{subsec:miniViT}
As shown in~\cref{fig:main_pipeline}, DPP predicts a set of non-uniform disparity planes ($D=\{D_{-L/2}, ... , D_{L/2}\}$) from the same input ($x_0$) to \textit{adaptively} represent $\lowrank$, on a Tensor Display~\cite{TD_layer}.
Having a \textit{disparity-aware} $TD$ layer instead of a uniformly distributed layer is beneficial for geometric accuracy and realism since objects can be present at non-uniform distances from each other~\cite{Li_VMPI}. 
We adopt a mini-ViT~\cite{AdaBinsBhat} to regress these values from image inputs similar to Mono~\cite{eccv_22_monoLFVR}.
It is imperative to note that our disparity plane prediction transformer has to solve the challenging task of estimating scene disparity as well by using \textit{just} the dual pixel input ($\{I_B, dp_L, dp_R\}$) unlike Mono that passes in scene depth as input.
\begin{figure*}[ht]
    \centering
    \includegraphics[width=\textwidth]{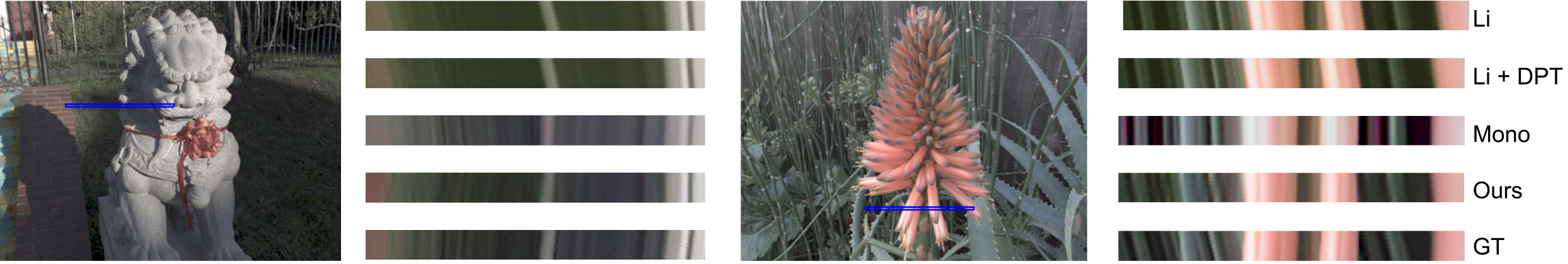}
    \caption{\textbf{Light Field EPI Comparison.} Our reconstructions are geometrically consistent, even for much higher spatial resolutions of $480 \times 640$ in zero shot settings. See~\cref{fig:quan_eval_resAgnos_and_dp_benefits}.}
    \label{fig:quaL_epi_geom}
\end{figure*}
\begin{table*}[ht]
    \centering
    \caption{\textbf{Zero Shot Cross Dataset Transfer.} Li~\cite{Li_VMPI}, Li+DPT and Mono-R~\cite{eccv_22_monoLFVR} are fine-tuned and do not operate under zero-shot settings. \bfb{Blue} and \bfg{green} are top two ZS methods. \textbf{Bold} is overall best, ZS or not. 
    Our method performs almost at par with Mono-R but with $13\times$ less parameters and $\sim100\%$ faster, see \cref{tab:temp_params_speed}}
    
    \begin{tabular}{llcccccccc|cc}
    \toprule
    \multirow{2}{*}{Res.} & \multirow{2}{*}{Model} & \multicolumn{2}{c}{TAMULF} & \multicolumn{2}{c}{Kalantari} & \multicolumn{2}{c}{Hybrid} & \multicolumn{2}{c}{Stanford} & \multicolumn{2}{c}{Average $\uparrow$}\\     
    
    \cmidrule(rl){3-4} \cmidrule(rl){5-6} \cmidrule(rl){7-8} \cmidrule(rl){9-10} \cmidrule(rl){11-12}
    
    & & PSNR & SSIM & PSNR & SSIM & PSNR & SSIM & PSNR & SSIM & PSNR & SSIM \\  
     
    \cmidrule(rl){1-12}

    \multirow{6}{*}{\rotatebox[origin=c]{90}{Lytro ($384\times 528$)}}
    
    & Niklaus & 12.56 & 0.75 & 16.69  & 0.83  & 19.14  & 0.85  & 17.84 & 0.83 & 16.55 & 0.81  \\    
  
    & Li & 22.23 & 0.85 & 26.14 & 0.89 & 28.67 & 0.91 & 26.81 & 0.89 & 25.96 & 0.88 \\ 
    
    & Li$+$DPT & 22.17 & 0.84 & 26.14  &  0.89 & 28.70  &  0.90 & 26.65 & 0.90 & 25.91 & 0.88 \\   
    
    & Mono & \bfb{23.05} & \bfb{0.88} &  \bfg{26.55} & \bfg{0.91} & \bfb{30.53}  & \bfg{0.94} & \bfg{27.87} & \bfg{0.92} & \bfg{27.00} & \bfg{0.91}  \\  
    
    & Mono$+$R & \textbf{24.34} & 0.88 &\textbf{26.81}  & \textbf{0.92}  & \textbf{30.88}  & 0.94 & 28.20  & 0.93 & \textbf{27.55} & \textbf{0.91} \\     
    
    \cmidrule(rl){2-12}
    
    & \textbf{Ours} & \bfg{22.75} & \bfg{0.85} & \bfb{26.73}  & \bfb{0.91} & \bfg{30.50} &  \bfb{0.94} & \bfb{28.54} & \bfb{0.93} & \bfb{27.13} & \bfb{0.91} \\          

    

    
    \bottomrule
    \end{tabular}
    \label{tab-1:ZSCDtransfer}
\end{table*}

\begin{figure*}[ht]
    \centering
    \includegraphics[width=0.99\textwidth]{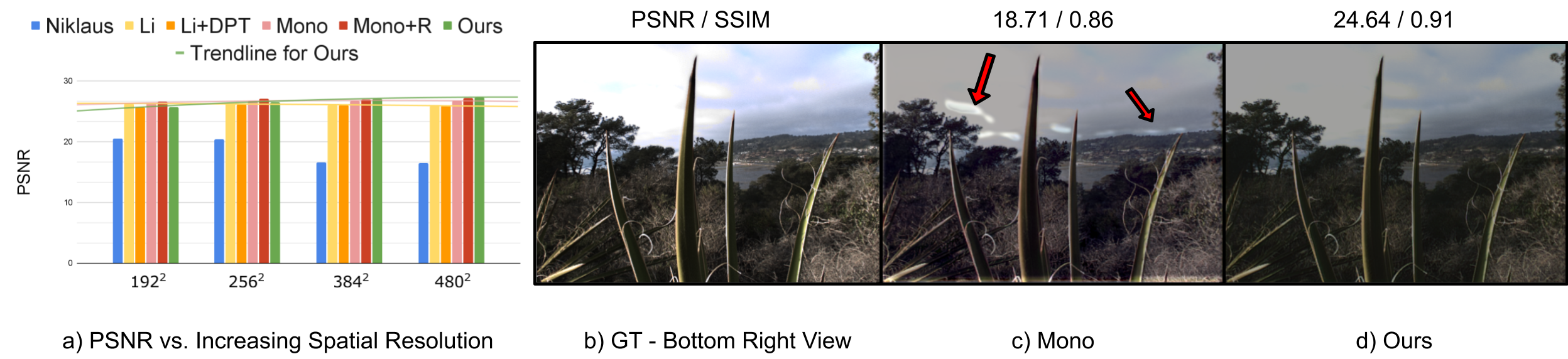}
    \caption{\textbf{Implicit 3D scene understanding essential for geometric consistency} comes from dark knowledge distillation. 
    a) PSNR trendline plateaus for others due to depth-input degradations from external networks (b - d), thus ceiling their reconstruction quality. Our independent method consistently rises with higher resolution.
    }
    \label{fig:quan_eval_resAgnos_and_dp_benefits}
\end{figure*}

\subsection{Structure Imposition: Adaptive Tensor Display Layer.}
After successfully regressing the disparity plane centers ($D$) and decoding $x_3$ to $\lowrank$ parallelly, $\lowrank$ is displayed as a structured light field ($\hat{L}$) on the tensor display layer using $D$, similar to~\cite{Selfvi}.
The operation of $TD(\cdot)$ can be formally described as follows:

\begin{equation}
\label{eq:td_eqn}
    \hat{L}(x,y,u,v) = TD(\lowrank, D) = \sum_{m=1}^M\prod_{l=-{L/2}}^{{L/2}} f_{m}^l(x + D_l u,y+ D_l v)
\end{equation}

where, $\hat{L}(x,y,u,v)$ represents the 4D light field and $(x,y)$ represents the spatial and $(u,v)$ the angular dimensions respectively.
This linear layer has no optimizable parameters. 

\subsection{Losses}
\label{subsec:losses}
In this section, we describe all the reconstruction losses used.

\textbf{Bins Chamfer Distance.}
Encourages the disparity planes ($D$) to be closer to the computed ground truth disparity~\cite{AdaBinsBhat} by minimizing:
\begin{equation}
    L_{bin} = \sum_{x \in X} \min_{y \in c(b)} \|x - y \|^2 + \sum_{y \in c(b)} \min_{x \in X} \|x - y \|^2.
    \label{eq:bin}
\end{equation}

where, $X$ is the set of centers of disparity planes from the computed ground truth disparity ($d_{st}$) and $c(b)$ is the set of centers of the predicted disparity planes ($D$). \\     

\textbf{Dark Geometric Consistency.}
Computes the error between warped SAIs to the central input-view $I_B$ based on a disparity estimate $d_{st}$ similar to~\cite{Selfvi}. 
$d_{st}$ is the \textit{dark knowledge} computed from the synthetic stereo teacher Unimatch~\cite{xu2023unifying_unimatch} using an additional view ($I_A$). 
Note that $I_A$ is \textit{never} an input to the student.
Warping of all angular views ($\mathbf{u}$) to the central view $\estcurrlf(\mathbf{0})$ is given by:
\begin{equation}
    \estcurrlf(\mathbf{u} \rightarrow \mathbf{0}) = \mathcal{W}\left(\estcurrlf\left(\mathbf{u}\rightarrow \mathbf{0}\right); \left(\mathbf{u} \rightarrow \mathbf{0}\right)ad_{st} \right),
    \label{eq:warping}
\end{equation}

where, $\mathcal W$ is the  bilinear inverse warping operator~\cite{invWarpOperator_NIPS}.
Then the geometric consistency loss is calculated as follows:
\begin{equation}
    L_{geo} = \sum_{\mathbf{u}} \| \estcurrlf\left(\mathbf{u} \rightarrow \mathbf{0}\right) - I_B \|_1.
    \label{eq:geo}
\end{equation}      
The baseline between $I_A$ and $I_B$ restricts the baseline of our LFs.
However, we introduce a hyperparameter $a \in [1,3]$, that scales the \textit{dark knowledge} $d_{st}$, to provide baseline control. 

\textbf{Dark Temporal Consistency.}
We integrate a flow teacher $\flownet$ based on RAFT~\cite{raft_RAFT} after a fixed number of iterations, completing the teacher ensemble.
$\flownet$ uses an additional video sequence frame as input ($I^{t+1}_B$) to compute optical flow $o_t = \flownet(I_B^t, I_B^{t+1})$. 
Note that $I^{t+1}_B$ is \textit{never} an input to the student.
The loss is enforced by warping all SAIs to central view $\estcurrlf\left(\mathbf{u}\rightarrow \mathbf{0}\right)$ (\cref{eq:warping}) and then
re-warping them to $I_B^{t+1}$ using $o_t$, similar to~\cite{Selfvi}:
\begin{equation}
    \loss_{temp} = \sum_{\mathbf{u}} 
 \| \warp\left(\warp\left(\estcurrlf\left(\mathbf{u} \rightarrow \mathbf{0} ; ad_{st}\right) \right) \rightarrow I^B_{t+1} ; o_{t}\right) - I^B_{t+1} \|_1.
    \label{eq:temp-loss}
\end{equation}      
\textbf{Photometric Loss} ensures faithful reconstruction of the central SAI ($\hat{L}(\mathbf{0})$) by minimizing $L_{photo} = \| \estcurrlf(\mathbf{0}) - \ensuremath{I_B}\|_1$. Where $I_B$ is the input.

\subsection{Implementation Details}
We use PyTorch~\cite{pytorch} and optimize using AdamW~\cite{adamW} with $1e-4$ LR and $0.0005$ weight decay. 
OneCycleLR scheduler \cite{scheduler} increases the LF for the first $10\%$ iterations and then cosine-anneals with a factor of $25$. 
Our batch size of 1 runs for 100 epochs with the geometry teacher~\cite{xu2023unifying_unimatch}, followed by 20 additional epochs with the flow teacher~\cite{raft_RAFT}. 3-skipped frames are used to capture disocclusion better. 
$7 \times 7$ is the LF output angular resolution.
We train on $256\times 192$ spatial resolution.
Hyperparameters: $\lambda_{photo}=1$, $\lambda_{geo}=1$, $\lambda_{bin}=2$, $\lambda_{temp}=0.2$, and $a=1.2$ are found empirically. We use one 12GB NVIDIA Titan X. 
\begin{table*}[ht]
    \centering
    \caption{\textbf{Best Temporal Consistency and Inference Speed.} \bfb{Blue} and \bfg{green} are top two. \textit{Left} shows temporal consistency metrics. (w/o $\flownet$) is our method trained without the flow teacher. \textit{Right} shows parameters and inference times.}
    \begin{tabular}{lccccc||cc}
    \toprule
    Models & Hybrid & Kalantari & TAMULF & Stanford & Average $\downarrow$ & Parameters (M) $\downarrow$ & $W_{GPU}$ (ms) $\downarrow$ \\      
    \midrule
    Niklaus & 0.357 & 0.070  & 0.219 & 0.065 & 0.177 & 153.23 & 667\\
    Li+\cite{deeplens2018} & 0.108 & 0.019 & 0.034 & 0.009 & 0.043 & \bfb{20.11} & \bfg{163} \\   
    Li+DPT & 0.108 & 0.016 & 0.033 & 0.008 & 0.042 & 351.61 & 185\\
    Mono & 0.103 & 0.017 & \bfg{0.028} & 0.006 & 0.038 & 481.19 & 184 \\  
    Mono-R & 0.102  & \bfg{0.016} & \bfb{0.027}  & \bfg{0.006} & 0.038 & 501.47  & 332 \\   
    \midrule
    Ours (w/o $\flownet$) & \bfg{0.069} & 0.017 & 0.033 & 0.019 & \bfg{0.034} & 38.18 & 159 \\  
    \textbf{Ours} & \bfb{0.037} & \bfb{0.009} & 0.055 & \bfb{0.004} & \bfb{0.026} & \bfg{38.18} & \bfb{159}\\ 
    \bottomrule
    \end{tabular}
    \label{tab:temp_params_speed}
\end{table*}

\section{Experiments}
\label{sec:experiments}
In this section, we extensively test, ablate, and compare our method \textit{trained exclusively on \textit{dp}MV} with existing state-of-the-art monocular LF reconstruction methods. 
Since \textit{dp}-channels are not available in ground truth LF datasets, we simulate them (procedure in supplementary).

\subsection{Comparison with State-of-the-Art}

\textbf{Datasets, Metrics, and Baselines} 
\textit{Datasets} used are: \textit{TAMULF}~\cite{Li_VMPI}, \textit{Hybrid}~\cite{wang2017light}, \textit{Kalantari}~\cite{LearningViewSynthesis_SIG16} and \textit{Stanford}~\cite{liff_stanford}.
To ensure fairness, all dataset splits, the angular resolution of GT LFs ($7\times 7$), homographic procedures for video LFs, and \textit{metrics}: PSNR, SSIM~\cite{ssim}, and RAFT~\cite{raft_RAFT} temporal consistency error are \textit{exactly} as that used by Mono~\cite{eccv_22_monoLFVR}.
Our \textit{baselines} include recent single-view LF reconstruction methods: Mono~\cite{eccv_22_monoLFVR} (Mono-R is the fully supervised refinement variant), Li~\cite{Li_VMPI} (Li+DPT variant uses depth from~\cite{Ranftl2020}) and zero-shot method, like ours, by Niklaus~\etal~\cite{Niklaus_Ken_Burns}.
We could not evaluate against~\cite{5d_monoVideo_recons_2019, ivan2019synthesizing, LFGAN_2023} since the source code was not available.

\textbf{Zero-Shot (ZS) Properties}

A) \textit{ZS cross-dataset transfer.} 
Our main fidelity comparison, quantitatively shown in \cref{tab-1:ZSCDtransfer} and qualitatively using epipolar images (EPIs) in \cref{fig:quaL_epi_geom}, is done under zero shot settings to show \textit{generalizability}. 
Unlike CNN-based methods, our approach does not require extensive fine-tuning. 
We outperform Li, Li+DPT across all datasets (\cref{tab-1:ZSCDtransfer}).

B) \textit{ZS higher spatial resolution synthesis} poses challenges for depth-dependent networks due to fidelity drops in external monocular depth estimators. 
Standard Definition resolution (480 $\times$ 640) results in a $\sim3\%$ decrease in DPT's performance~\cite{Ranftl2020}, leading to beveled pixels along difficult depth-edges in SAIs (\cref{fig:quan_eval_resAgnos_and_dp_benefits} b-d).
The advantage of dark knowledge distillation and dual pixels' fine-grained guidance in foreground-background separation edges is most prominent in these settings. 
Quantitatively, our PSNR rises while other depth-dependent methods plateau due to severe degradations (\cref{fig:quan_eval_resAgnos_and_dp_benefits} a).


\textbf{Temporal Consistency.} 
We use the RAFT~\cite{raft_RAFT} based temporal loss operation described in~\cref{subsec:losses} ($L_{temp}$) for evaluation. 
Our LF videos achieve the highest consistency. See~\cref{tab:temp_params_speed}. See supplementary for LF videos.

\textbf{Parameters and Inference Speed.}
Shown in~\cref{tab:temp_params_speed}.
Our model achieves competitive fidelity using $13\times$ lesser parameters and at $2\times$ inference speed compared to Mono-R.  
The reported numbers are averaged across all test sets.

\subsection{Ablations and Hyper-parameter Study}
\begin{table}[h]
    \centering
    \caption{\textbf{Input Ablations}. Cvg-Ep stands for convergence epoch decided by $L_{geo}$'s plateau threshold.
    $+d_{st}$ variants converge $L_{geo}$ before $L_{photo}$ due to explicit disparity as input.}
    \begin{tabular}{lccc}
    \toprule
    Input-to-Model & Cvg-Ep & PSNR $\uparrow$ & SSIM $\uparrow$ \\  
    \cmidrule{1-4}
    $I_B$ & \xmark & - & -  \\
    $I_B+d_{st}$ & \textbf{$<$30} & 24.871 & 0.90 \\
    $I_B+dp+d_{st}$ & $\sim$30 & 24.717 & 0.89 \\
    \midrule
    $I_B+dp$ (\textbf{Proposed}) & $<$80 & \textbf{26.865} & \textbf{0.91}   \\  
    \bottomrule
\end{tabular}
\label{tab:input_ablation}
\end{table}

\begin{table}[h]
\centering
\caption{\textbf{Loss Ablations} $L_{geo}$ (\cref{eq:geo}) and bins $L_{bin}$ (\cref{eq:bin}) are the most critical. Metrics are averaged across all GT-test datasets}
\begin{tabular}{lccccc}
    \toprule
    Ver. & $L_{geo}$ & $L_{pho}$ & $L_{bin}$ & PSNR & SSIM \\   
    \cmidrule{1-6}
    V1 & \checkmark & \xmark & \xmark & 22.89 & 0.85 \\ 
    V2 & \xmark & \checkmark & \xmark & 13.01 & 0.59 \\ 
    V3 & \xmark & \xmark & \checkmark & 4.34 & 0.50 \\  
    V4 & \xmark & \checkmark & \checkmark & 23.80 & 0.85  \\
    V5 & \checkmark & \xmark & \checkmark & 24.76 & 0.86  \\
    V6 & \checkmark & \checkmark & \xmark & 23.39 & 0.86 \\
    \midrule
    \textbf{Ours} & \checkmark & \checkmark & \checkmark & \textbf{25.39} & \textbf{0.89}  \\
    \bottomrule
\end{tabular}
\label{tab-7:loss_ablations}
\end{table}

\textbf{Input Ablation Breakdown}
In \cref{tab:input_ablation}, when explicit disparity ($d_{st}$ computed by Unimatch) is given as input instead of implicit distillation, $L_{geo}$ converges first followed by $L_{photo}$.
This suggests a limited internal 3D scene understanding, as the model heavily relies on $d_{st}$. 
When $d_{st}$'s quality decreases due to external network limitations, the performance of the explicit input variant also suffers.
Mono faces similar issues (\cref{fig:quan_eval_resAgnos_and_dp_benefits}).
Also, our network's representational power is insufficient for the monocular variant to converge.

\begin{figure}[h]
    \centering
    \includegraphics[width=0.95\columnwidth]{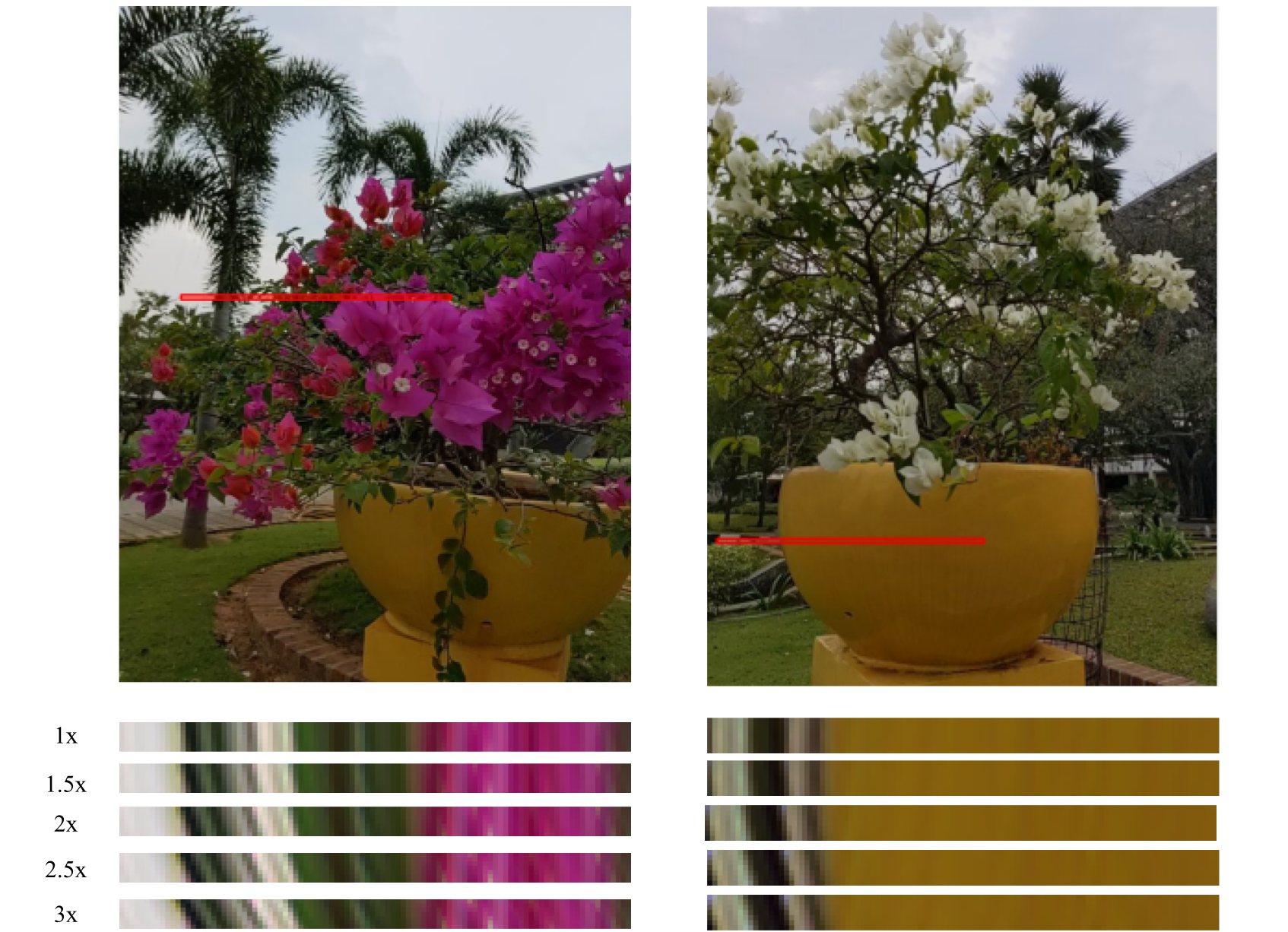}
    \caption{\textbf{Controllable Baseline} using dark knowledge scaling hyperparameter ($a$).}
    \label{fig:real_vary_baselines}
\end{figure}

\textbf{Dark Knowledge Scaling Hyper-parameter for Wider LF Baselines}
$a > 1$ enables higher baseline LF reconstruction ($d_{wide} \sim ad_{st}$). 
Qualitative EPI slices of reconstructed \textit{dp}MV LFs are shown in~\cref{fig:real_vary_baselines}.

\subsection{XR Applications on LF Video Reconstructions}
\begin{figure}[h]
        \centering
        \includegraphics[width=\columnwidth]{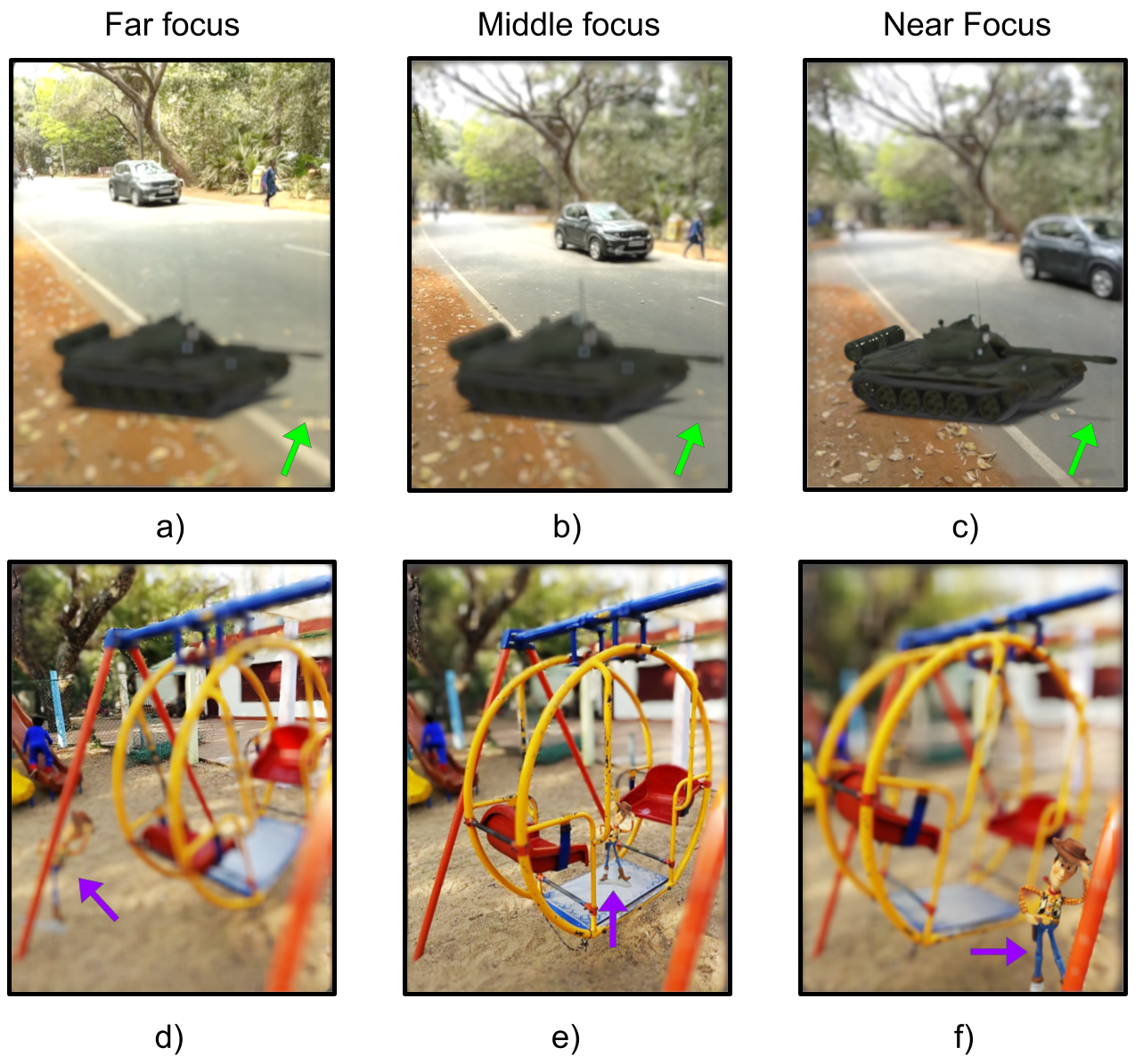}
        \caption{\textbf{XR Applications-1 of Reconstructed LF videos} a)-c) \bfg{Green} arrows demonstrate correct shadows applied on virtual objects. d)-f) \textbf{\textcolor{violet}{Purple}} shows occlusion-aware and consistent insertion of animated characters in a real LF video}
    \label{fig:last-app_vary_baseline}
\end{figure}

\begin{figure}[h]
    \centering
    \includegraphics[width=0.9\columnwidth]{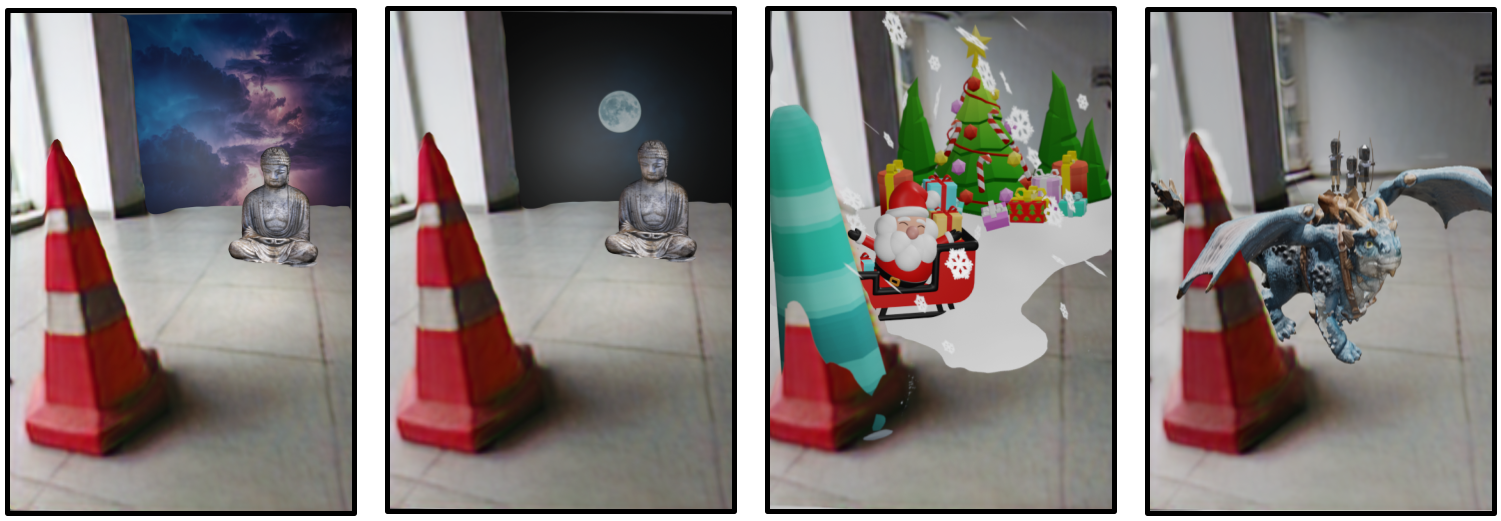}
    \caption{\textbf{XR Applications-2.} Virtual object insertion, depth-aware background editing, and recoloring on our reconstructed light fields.}
    \label{fig:xr_applications}
\end{figure}

Inspired by simple shift-and-add SAI procedures for refocusing and synthetic aperture control by~\cite{ng2005light}, and edge-aware XR by~\cite{khan2021edgeaware}, we showcase simultaneous refocusing, aperture control, and virtual object insertion on reconstructed \textit{dp}MV LF videos in~\cref{fig:last-app_vary_baseline} and~\cref{fig:xr_applications}.

\section{Discussion}
\label{sec:discuss}
We introduced the largest dp-video dataset with computed disparity estimates alongside a dark knowledge distillation hypothesis that enables 3D scene understanding using a single \textit{dp}-view. 
We demonstrated the fastest and most temporally consistent light field reconstruction using our dataset and distillation. 
We demonstrate zero-shot generalization with vision transformers for practical dp-smartphone 3D reconstruction. 
\textit{Limitations} include adaptation for night-time scenes, challenges with dark knowledge scaling ($a > 3$), and reconstruction of transmissive and reflective surfaces. 

\bibliographystyle{IEEEtran}
\bibliography{references}

\begin{thebibliography}{10}
\providecommand{\url}[1]{#1}
\csname url@samestyle\endcsname
\providecommand{\newblock}{\relax}
\providecommand{\bibinfo}[2]{#2}
\providecommand{\BIBentrySTDinterwordspacing}{\spaceskip=0pt\relax}
\providecommand{\BIBentryALTinterwordstretchfactor}{4}
\providecommand{\BIBentryALTinterwordspacing}{\spaceskip=\fontdimen2\font plus
\BIBentryALTinterwordstretchfactor\fontdimen3\font minus \fontdimen4\font\relax}
\providecommand{\BIBforeignlanguage}[2]{{%
\expandafter\ifx\csname l@#1\endcsname\relax
\typeout{** WARNING: IEEEtran.bst: No hyphenation pattern has been}%
\typeout{** loaded for the language `#1'. Using the pattern for}%
\typeout{** the default language instead.}%
\else
\language=\csname l@#1\endcsname
\fi
#2}}
\providecommand{\BIBdecl}{\relax}
\BIBdecl

\bibitem{autofocus}
A.~Abuolaim and M.~Brown, ``Online lens motion smoothing for video autofocus,'' in \emph{Proceedings of the IEEE/CVF Winter Conference on Applications of Computer Vision (WACV)}, March 2020.

\bibitem{GargDualPixelsICCV2019_DPNet}
R.~Garg, N.~Wadhwa, S.~Ansari, and J.~T. Barron, ``Learning single camera depth estimation using dual-pixels,'' \emph{ICCV}, 2019.

\bibitem{Pan_2021_CVPR_DDDNet_DP_Disp}
L.~Pan, S.~Chowdhury, R.~Hartley, M.~Liu, H.~Zhang, and H.~Li, ``Dual pixel exploration: Simultaneous depth estimation and image restoration,'' in \emph{Proceedings of the IEEE/CVF Conference on Computer Vision and Pattern Recognition (CVPR)}, June 2021, pp. 4340--4349.

\bibitem{punnappurath2020modeling}
A.~Punnappurath, A.~Abuolaim, M.~Afifi, and M.~S. Brown, ``Modeling defocus-disparity in dual-pixel sensors,'' in \emph{IEEE International Conference on Computational Photography (ICCP)}, 2020.

\bibitem{Xin_2021_ICCV_dual_pixel_cmu}
S.~Xin, N.~Wadhwa, T.~Xue, J.~T. Barron, P.~P. Srinivasan, J.~Chen, I.~Gkioulekas, and R.~Garg, ``Defocus map estimation and deblurring from a single dual-pixel image,'' \emph{IEEE/CVF International Conference on Computer Vision (ICCV)}, 2021.

\bibitem{refocus_wadwa_18}
\BIBentryALTinterwordspacing
N.~Wadhwa, R.~Garg, D.~E. Jacobs, B.~E. Feldman, N.~Kanazawa, R.~Carroll, Y.~Movshovitz-Attias, J.~T. Barron, Y.~Pritch, and M.~Levoy, ``Synthetic depth-of-field with a single-camera mobile phone,'' \emph{ACM Transactions on Graphics}, vol.~37, no.~4, p. 1–13, Jul. 2018. [Online]. Available: \url{http://dx.doi.org/10.1145/3197517.3201329}
\BIBentrySTDinterwordspacing

\bibitem{kitti_dataset}
M.~Menze and A.~Geiger, ``Object scene flow for autonomous vehicles,'' in \emph{Conference on Computer Vision and Pattern Recognition (CVPR)}, 2015.

\bibitem{eth_dataset_stereo}
T.~Sch\"ops, J.~L. Sch\"onberger, S.~Galliani, T.~Sattler, K.~Schindler, M.~Pollefeys, and A.~Geiger, ``A multi-view stereo benchmark with high-resolution images and multi-camera videos,'' in \emph{Conference on Computer Vision and Pattern Recognition (CVPR)}, 2017.

\bibitem{middlebury}
\BIBentryALTinterwordspacing
D.~Scharstein, H.~Hirschm{\"u}ller, Y.~Kitajima, G.~Krathwohl, N.~Nesic, X.~Wang, and P.~Westling, ``High-resolution stereo datasets with subpixel-accurate ground truth,'' in \emph{German Conference on Pattern Recognition}, 2014. [Online]. Available: \url{https://api.semanticscholar.org/CorpusID:14915763}
\BIBentrySTDinterwordspacing

\bibitem{colmap_COLMAP}
J.~L. Sch{\"o}nberger, E.~Zheng, J.-M. Frahm, and M.~Pollefeys, ``Pixelwise view selection for unstructured multi-view stereo,'' in \emph{Computer Vision -- ECCV 2016}, B.~Leibe, J.~Matas, N.~Sebe, and M.~Welling, Eds.\hskip 1em plus 0.5em minus 0.4em\relax Cham: Springer International Publishing, 2016, pp. 501--518.

\bibitem{sceneflow_dataset}
\BIBentryALTinterwordspacing
N.~Mayer, E.~Ilg, P.~H{\"a}usser, P.~Fischer, D.~Cremers, A.~Dosovitskiy, and T.~Brox, ``A large dataset to train convolutional networks for disparity, optical flow, and scene flow estimation,'' in \emph{IEEE International Conference on Computer Vision and Pattern Recognition (CVPR)}, 2016, arXiv:1512.02134. [Online]. Available: \url{http://lmb.informatik.uni-freiburg.de/Publications/2016/MIFDB16}
\BIBentrySTDinterwordspacing

\bibitem{xu2023unifying_unimatch}
H.~Xu, J.~Zhang, J.~Cai, H.~Rezatofighi, F.~Yu, D.~Tao, and A.~Geiger, ``Unifying flow, stereo and depth estimation,'' \emph{IEEE Transactions on Pattern Analysis and Machine Intelligence}, 2023.

\bibitem{kd_first_01}
\BIBentryALTinterwordspacing
C.~Buciluundefined, R.~Caruana, and A.~Niculescu-Mizil, ``Model compression,'' in \emph{Proceedings of the 12th ACM SIGKDD International Conference on Knowledge Discovery and Data Mining}, ser. KDD '06.\hskip 1em plus 0.5em minus 0.4em\relax New York, NY, USA: Association for Computing Machinery, 2006, p. 535–541. [Online]. Available: \url{https://doi.org/10.1145/1150402.1150464}
\BIBentrySTDinterwordspacing

\bibitem{kd_survey_02}
\BIBentryALTinterwordspacing
L.~Wang and K.-J. Yoon, ``Knowledge distillation and student-teacher learning for visual intelligence: A review and new outlooks,'' \emph{IEEE Transactions on Pattern Analysis and Machine Intelligence}, vol.~44, no.~6, p. 3048–3068, Jun. 2022. [Online]. Available: \url{http://dx.doi.org/10.1109/TPAMI.2021.3055564}
\BIBentrySTDinterwordspacing

\bibitem{depth_anything}
L.~Yang, B.~Kang, Z.~Huang, X.~Xu, J.~Feng, and H.~Zhao, ``Depth anything: Unleashing the power of large-scale unlabeled data,'' in \emph{CVPR}, 2024.

\bibitem{midas}
R.~Ranftl, K.~Lasinger, D.~Hafner, K.~Schindler, and V.~Koltun, ``Towards robust monocular depth estimation: Mixing datasets for zero-shot cross-dataset transfer,'' \emph{IEEE Transactions on Pattern Analysis and Machine Intelligence (TPAMI)}, 2020.

\bibitem{Ranftl2020}
R.~Ranftl, A.~Bochkovskiy, and V.~Koltun, ``Vision transformers for dense prediction,'' \emph{ArXiv preprint}, 2021.

\bibitem{MarcLevoy_LFRendering}
\BIBentryALTinterwordspacing
M.~Levoy and P.~Hanrahan, \emph{Light Field Rendering}, 1st~ed.\hskip 1em plus 0.5em minus 0.4em\relax New York, NY, USA: Association for Computing Machinery, 2023. [Online]. Available: \url{https://doi.org/10.1145/3596711.3596759}
\BIBentrySTDinterwordspacing

\bibitem{khan2021edgeaware}
N.~Khan, M.~H. Kim, and J.~Tompkin, ``Edge-aware bidirectional diffusion for dense depth estimation from light fields,'' \emph{British Machine Vision Conference}, 2021.

\bibitem{khan2020vclfd}
J.~T. Numair~Khan, Min H.~Kim, ``View-consistent {4D} lightfield depth estimation,'' \emph{British Machine Vision Conference}, 2020.

\bibitem{ng2005light}
R.~Ng, M.~Levoy, M.~Br{\'e}dif, G.~Duval, M.~Horowitz, and P.~Hanrahan, ``Light field photography with a hand-held plenoptic camera,'' Ph.D. dissertation, Stanford University, 2005.

\bibitem{lowCostMultiCameraArray}
\BIBentryALTinterwordspacing
M.~Broxton, J.~Busch, J.~Dourgarian, M.~DuVall, D.~Erickson, D.~Evangelakos, J.~Flynn, R.~Overbeck, M.~Whalen, and P.~Debevec, ``A low cost multi-camera array for panoramic light field video capture,'' in \emph{SIGGRAPH Asia 2019 Posters}, ser. SA '19.\hskip 1em plus 0.5em minus 0.4em\relax New York, NY, USA: Association for Computing Machinery, 2019. [Online]. Available: \url{https://doi.org/10.1145/3355056.3364593}
\BIBentrySTDinterwordspacing

\bibitem{systemForAcquiringLF_google}
\BIBentryALTinterwordspacing
R.~S. Overbeck, D.~Erickson, D.~Evangelakos, M.~Pharr, and P.~Debevec, ``A system for acquiring, processing, and rendering panoramic light field stills for virtual reality,'' \emph{ACM Trans. Graph.}, vol.~37, no.~6, dec 2018. [Online]. Available: \url{https://doi.org/10.1145/3272127.3275031}
\BIBentrySTDinterwordspacing

\bibitem{broxton2020immersive_google}
M.~Broxton, J.~Flynn, R.~Overbeck, D.~Erickson, P.~Hedman, M.~DuVall, J.~Dourgarian, J.~Busch, M.~Whalen, and P.~Debevec, ``Immersive light field video with a layered mesh representation,'' vol.~39, no.~4, pp. 86:1--86:15, 2020.

\bibitem{eccv_22_monoLFVR}
\BIBentryALTinterwordspacing
S.~Govindarajan, P.~Shedligeri, Sarah, and K.~Mitra, ``Synthesizing light field video from monocular video.''\hskip 1em plus 0.5em minus 0.4em\relax Berlin, Heidelberg: Springer-Verlag, 2022, p. 162–180. [Online]. Available: \url{https://doi.org/10.1007/978-3-031-20071-7_10}
\BIBentrySTDinterwordspacing

\bibitem{Li_VMPI}
Q.~Li and N.~Khademi~Kalantari, ``Synthesizing light field from a single image with variable mpi and two network fusion,'' \emph{ACM Transactions on Graphics}, vol.~39, no.~6, 12 2020.

\bibitem{Srinivasan_2017_ICCV_4D_RGBD}
P.~P. Srinivasan, T.~Wang, A.~Sreelal, R.~Ramamoorthi, and R.~Ng, ``Learning to synthesize a 4d rgbd light field from a single image,'' in \emph{Proceedings of the IEEE International Conference on Computer Vision (ICCV)}, Oct 2017.

\bibitem{5d_monoVideo_recons_2019}
K.~Bae, A.~Ivan, H.~Nagahara, and I.~K. Park, ``5d light field synthesis from a monocular video,'' 2019.

\bibitem{ivan2019synthesizing}
A.~Ivan, I.~K. Park \emph{et~al.}, ``Synthesizing a 4d spatio-angular consistent light field from a single image,'' \emph{arXiv preprint arXiv:1903.12364}, 2019.

\bibitem{LFGAN_2023}
\BIBentryALTinterwordspacing
B.~Chen, L.~Ruan, and M.-L. Lam, ``Lfgan: 4d light field synthesis from a single rgb image,'' \emph{ACM Trans. Multimedia Comput. Commun. Appl.}, vol.~16, no.~1, feb 2020. [Online]. Available: \url{https://doi.org/10.1145/3366371}
\BIBentrySTDinterwordspacing

\bibitem{Selfvi}
P.~Shedligeri, F.~Schiffers, S.~Ghosh, O.~Cossairt, and K.~Mitra, ``Selfvi: Self-supervised light-field video reconstruction from stereo video,'' in \emph{Proceedings of the IEEE/CVF International Conference on Computer Vision (ICCV)}, October 2021, pp. 2491--2501.

\bibitem{LearningViewSynthesis_SIG16}
N.~K. Kalantari, T.-C. Wang, and R.~Ramamoorthi, ``Learning-based view synthesis for light field cameras,'' \emph{ACM Transactions on Graphics (Proceedings of SIGGRAPH Asia 2016)}, vol.~35, no.~6, 2016.

\bibitem{Niklaus_Ken_Burns}
S.~Niklaus, L.~Mai, J.~Yang, and F.~Liu, ``3d ken burns effect from a single image,'' \emph{ACM Transactions on Graphics}, vol.~38, no.~6, pp. 184:1--184:15, 2019.

\bibitem{rebuttal_01}
\BIBentryALTinterwordspacing
A.~Jaiswal, A.~R. Babu, M.~Z. Zadeh, D.~Banerjee, and F.~Makedon, ``A survey on contrastive self-supervised learning,'' \emph{Technologies}, vol.~9, no.~1, 2021. [Online]. Available: \url{https://www.mdpi.com/2227-7080/9/1/2}
\BIBentrySTDinterwordspacing

\bibitem{rebuttal_02}
V.~Rani, S.~Nabi, M.~Kumar, A.~Mittal, and K.~Saluja, ``Self-supervised learning: A succinct review,'' \emph{Archives of Computational Methods in Engineering}, vol.~30, 01 2023.

\bibitem{dosovitskiy2020vit}
A.~Dosovitskiy, L.~Beyer, A.~Kolesnikov, D.~Weissenborn, X.~Zhai, T.~Unterthiner, M.~Dehghani, M.~Minderer, G.~Heigold, S.~Gelly, J.~Uszkoreit, and N.~Houlsby, ``An image is worth 16x16 words: Transformers for image recognition at scale,'' \emph{ICLR}, 2021.

\bibitem{raft_RAFT}
\BIBentryALTinterwordspacing
Z.~Teed and J.~Deng, ``Raft: Recurrent all-pairs field transforms for optical flow,'' in \emph{Computer Vision – ECCV 2020: 16th European Conference, Glasgow, UK, August 23–28, 2020, Proceedings, Part II}.\hskip 1em plus 0.5em minus 0.4em\relax Berlin, Heidelberg: Springer-Verlag, 2020, p. 402–419. [Online]. Available: \url{https://doi.org/10.1007/978-3-030-58536-5_24}
\BIBentrySTDinterwordspacing

\bibitem{kd_first_02}
\BIBentryALTinterwordspacing
G.~E. Hinton, O.~Vinyals, and J.~Dean, ``Distilling the knowledge in a neural network,'' \emph{CoRR}, vol. abs/1503.02531, 2015. [Online]. Available: \url{http://arxiv.org/abs/1503.02531}
\BIBentrySTDinterwordspacing

\bibitem{do_nn_need_to_be_deep}
\BIBentryALTinterwordspacing
J.~Ba and R.~Caruana, ``Do deep nets really need to be deep?'' in \emph{Advances in Neural Information Processing Systems}, Z.~Ghahramani, M.~Welling, C.~Cortes, N.~Lawrence, and K.~Weinberger, Eds., vol.~27.\hskip 1em plus 0.5em minus 0.4em\relax Curran Associates, Inc., 2014. [Online]. Available: \url{https://proceedings.neurips.cc/paper_files/paper/2014/file/ea8fcd92d59581717e06eb187f10666d-Paper.pdf}
\BIBentrySTDinterwordspacing

\bibitem{kd_survey_01}
\BIBentryALTinterwordspacing
J.~Gou, B.~Yu, S.~J. Maybank, and D.~Tao, ``Knowledge distillation: A survey,'' \emph{International Journal of Computer Vision}, vol. 129, no.~6, p. 1789–1819, Mar. 2021. [Online]. Available: \url{http://dx.doi.org/10.1007/s11263-021-01453-z}
\BIBentrySTDinterwordspacing

\bibitem{response_01}
W.~Park, D.~Kim, Y.~Lu, and M.~Cho, ``Relational knowledge distillation,'' in \emph{Proceedings of the IEEE Conference on Computer Vision and Pattern Recognition}, 2019, pp. 3967--3976.

\bibitem{response_02}
F.~Tung and G.~Mori, ``Similarity-preserving knowledge distillation,'' in \emph{International Conference on Computer Vision (ICCV)}, 2019.

\bibitem{response_03}
B.~Peng, X.~Jin, J.~Liu, D.~Li, Y.~Wu, Y.~Liu, S.~Zhou, and Z.~Zhang, ``Correlation congruence for knowledge distillation,'' in \emph{Proceedings of the IEEE/CVF International Conference on Computer Vision (ICCV)}, October 2019.

\bibitem{response_04}
\BIBentryALTinterwordspacing
J.~Yim, D.~Joo, J.~Bae, and J.~Kim, ``A gift from knowledge distillation: Fast optimization, network minimization and transfer learning,'' in \emph{2017 IEEE Conference on Computer Vision and Pattern Recognition (CVPR)}.\hskip 1em plus 0.5em minus 0.4em\relax Los Alamitos, CA, USA: IEEE Computer Society, jul 2017, pp. 7130--7138. [Online]. Available: \url{https://doi.ieeecomputersociety.org/10.1109/CVPR.2017.754}
\BIBentrySTDinterwordspacing

\bibitem{feature_01}
\BIBentryALTinterwordspacing
A.~Romero, N.~Ballas, S.~E. Kahou, A.~Chassang, C.~Gatta, and Y.~Bengio, ``Fitnets: Hints for thin deep nets,'' in \emph{3rd International Conference on Learning Representations, {ICLR} 2015, San Diego, CA, USA, May 7-9, 2015, Conference Track Proceedings}, Y.~Bengio and Y.~LeCun, Eds., 2015. [Online]. Available: \url{http://arxiv.org/abs/1412.6550}
\BIBentrySTDinterwordspacing

\bibitem{feature_02}
\BIBentryALTinterwordspacing
S.~Zagoruyko and N.~Komodakis, ``Paying more attention to attention: Improving the performance of convolutional neural networks via attention transfer,'' in \emph{ICLR}, 2017. [Online]. Available: \url{https://arxiv.org/abs/1612.03928}
\BIBentrySTDinterwordspacing

\bibitem{contrastive_01}
Y.~Tian, D.~Krishnan, and P.~Isola, ``Contrastive representation distillation,'' in \emph{International Conference on Learning Representations}, 2020.

\bibitem{data_eff_transformers_deit}
H.~Touvron, M.~Cord, M.~Douze, F.~Massa, A.~Sablayrolles, and H.~Jégou, ``Training data-efficient image transformers and distillation through attention,'' 2021.

\bibitem{conVit_01}
\BIBentryALTinterwordspacing
F.~Wu, A.~Fan, A.~Baevski, Y.~Dauphin, and M.~Auli, ``Pay less attention with lightweight and dynamic convolutions,'' in \emph{International Conference on Learning Representations}, 2019. [Online]. Available: \url{https://arxiv.org/abs/1901.10430}
\BIBentrySTDinterwordspacing

\bibitem{conVit_02}
Z.~Peng, W.~Huang, S.~Gu, L.~Xie, Y.~Wang, J.~Jiao, and Q.~Ye, ``Conformer: Local features coupling global representations for visual recognition,'' in \emph{Proceedings of the IEEE/CVF International Conference on Computer Vision (ICCV)}, October 2021, pp. 367--376.

\bibitem{conVit_03}
Y.~Wang, Y.~Yang, J.~Bai, M.~Zhang, J.~Bai, J.~Yu, C.~Zhang, G.~Huang, and Y.~Tong, ``Evolving attention with residual convolutions,'' 2021.

\bibitem{conVit_04}
K.~Yuan, S.~Guo, Z.~Liu, A.~Zhou, F.~Yu, and W.~Wu, ``Incorporating convolution designs into visual transformers,'' in \emph{Proceedings of the IEEE/CVF International Conference on Computer Vision (ICCV)}, October 2021, pp. 579--588.

\bibitem{wu2021cvt}
H.~Wu, B.~Xiao, N.~Codella, M.~Liu, X.~Dai, L.~Yuan, and L.~Zhang, ``Cvt: Introducing convolutions to vision transformers,'' \emph{arXiv preprint arXiv:2103.15808}, 2021.

\bibitem{early_convolutions}
\BIBentryALTinterwordspacing
T.~Xiao, M.~Singh, E.~Mintun, T.~Darrell, P.~Dollar, and R.~Girshick, ``Early convolutions help transformers see better,'' in \emph{Advances in Neural Information Processing Systems}, M.~Ranzato, A.~Beygelzimer, Y.~Dauphin, P.~Liang, and J.~W. Vaughan, Eds., vol.~34.\hskip 1em plus 0.5em minus 0.4em\relax Curran Associates, Inc., 2021, pp. 30\,392--30\,400. [Online]. Available: \url{https://proceedings.neurips.cc/paper_files/paper/2021/file/ff1418e8cc993fe8abcfe3ce2003e5c5-Paper.pdf}
\BIBentrySTDinterwordspacing

\bibitem{token_to_vit}
L.~Yuan, Y.~Chen, T.~Wang, W.~Yu, Y.~Shi, Z.-H. Jiang, F.~E. Tay, J.~Feng, and S.~Yan, ``Tokens-to-token vit: Training vision transformers from scratch on imagenet,'' in \emph{Proceedings of the IEEE/CVF International Conference on Computer Vision (ICCV)}, October 2021, pp. 558--567.

\bibitem{vldi_Bak_2023}
J.~Bak and I.~K. Park, ``Light field synthesis from a monocular image using variable ldi,'' in \emph{Proceedings of the IEEE/CVF Conference on Computer Vision and Pattern Recognition (CVPR) Workshops}, June 2023, pp. 3398--3406.

\bibitem{bino-LF}
Z.~Zhang, Y.~Liu, and Q.~Dai, ``Light field from micro-baseline image pair,'' in \emph{2015 IEEE Conference on Computer Vision and Pattern Recognition (CVPR)}, 2015, pp. 3800--3809.

\bibitem{mildenhall2020nerf}
\BIBentryALTinterwordspacing
B.~Mildenhall, P.~P. Srinivasan, M.~Tancik, J.~T. Barron, R.~Ramamoorthi, and R.~Ng, ``Nerf: Representing scenes as neural radiance fields for view synthesis,'' in \emph{Proceedings of the European Conference on Computer Vision (ECCV)}, 2020. [Online]. Available: \url{http://arxiv.org/abs/2003.08934v2}
\BIBentrySTDinterwordspacing

\bibitem{liu2021neural}
L.~Liu, J.~Gu, K.~Z. Lin, T.-S. Chua, and C.~Theobalt, ``Neural sparse voxel fields,'' 2021.

\bibitem{zhang2020nerf}
K.~Zhang, G.~Riegler, N.~Snavely, and V.~Koltun, ``Nerf++: Analyzing and improving neural radiance fields,'' 2020.

\bibitem{Xu_2022_SinNeRF}
D.~Xu, Y.~Jiang, P.~Wang, Z.~Fan, H.~Shi, and Z.~Wang, ``Sinnerf: Training neural radiance fields on complex scenes from a single image,'' 2022.

\bibitem{yu2021pixelnerf}
A.~Yu, V.~Ye, M.~Tancik, and A.~Kanazawa, ``{pixelNeRF}: Neural radiance fields from one or few images,'' in \emph{CVPR}, 2021.

\bibitem{wu2017light}
G.~Wu, M.~Zhao, L.~Wang, Q.~Dai, T.~Chai, and Y.~Liu, ``Light field reconstruction using deep convolutional network on epi,'' in \emph{Proceedings of the IEEE Conference on Computer Vision and Pattern Recognition}, 2017, pp. 6319--6327.

\bibitem{wang2018end}
Y.~Wang, F.~Liu, Z.~Wang, G.~Hou, Z.~Sun, and T.~Tan, ``End-to-end view synthesis for light field imaging with pseudo 4dcnn,'' in \emph{Proceedings of the European Conference on Computer Vision (ECCV)}, 2018, pp. 333--348.

\bibitem{LocalLF_fusion_2020}
\BIBentryALTinterwordspacing
B.~Mildenhall, P.~P. Srinivasan, R.~Ortiz-Cayon, N.~K. Kalantari, R.~Ramamoorthi, R.~Ng, and A.~Kar, ``Local light field fusion: Practical view synthesis with prescriptive sampling guidelines,'' \emph{ACM Trans. Graph.}, vol.~38, no.~4, jul 2019. [Online]. Available: \url{https://doi.org/10.1145/3306346.3322980}
\BIBentrySTDinterwordspacing

\bibitem{flynn2019deepview}
J.~Flynn, M.~Broxton, P.~Debevec, M.~DuVall, G.~Fyffe, R.~Overbeck, N.~Snavely, and R.~Tucker, ``Deepview: View synthesis with learned gradient descent,'' 2019.

\bibitem{Bemana2020xfields}
M.~Bemana, K.~Myszkowski, H.-P. Seidel, and T.~Ritschel, ``X-fields: Implicit neural view-, light- and time-image interpolation,'' \emph{ACM Transactions on Graphics (Proc. SIGGRAPH Asia 2020)}, vol.~39, no.~6, 2020.

\bibitem{single_view_mpi}
R.~Tucker and N.~Snavely, ``Single-view view synthesis with multiplane images,'' in \emph{The IEEE Conference on Computer Vision and Pattern Recognition (CVPR)}, June 2020.

\bibitem{SMPI}
M.~Zhang, J.~Wang, X.~Li, Y.~Huang, Y.~Sato, and Y.~Lu, ``Structural multiplane image: Bridging neural view synthesis and 3d reconstruction,'' in \emph{Proceedings of the IEEE/CVF Conference on Computer Vision and Pattern Recognition}, 2023, pp. 16\,707--16\,716.

\bibitem{what_knowledge_ojha}
\BIBentryALTinterwordspacing
U.~Ojha, Y.~Li, A.~Sundara~Rajan, Y.~Liang, and Y.~J. Lee, ``What knowledge gets distilled in knowledge distillation?'' in \emph{Advances in Neural Information Processing Systems}, A.~Oh, T.~Neumann, A.~Globerson, K.~Saenko, M.~Hardt, and S.~Levine, Eds., vol.~36.\hskip 1em plus 0.5em minus 0.4em\relax Curran Associates, Inc., 2023, pp. 11\,037--11\,048. [Online]. Available: \url{https://proceedings.neurips.cc/paper_files/paper/2023/file/2433fec2144ccf5fea1c9c5ebdbc3924-Paper-Conference.pdf}
\BIBentrySTDinterwordspacing

\bibitem{student_better_than_teacher}
J.~Cho and B.~Hariharan, ``On the efficacy of knowledge distillation,'' 10 2019, pp. 4793--4801.

\bibitem{xu2022attention_ACVNet}
G.~Xu, J.~Cheng, P.~Guo, and X.~Yang, ``Attention concatenation volume for accurate and efficient stereo matching,'' in \emph{Proceedings of the IEEE/CVF Conference on Computer Vision and Pattern Recognition}, 2022, pp. 12\,981--12\,990.

\bibitem{STTR}
Z.~Li, X.~Liu, N.~Drenkow, A.~Ding, F.~X. Creighton, R.~H. Taylor, and M.~Unberath, ``Revisiting stereo depth estimation from a sequence-to-sequence perspective with transformers,'' in \emph{Proceedings of the IEEE/CVF International Conference on Computer Vision (ICCV)}, October 2021, pp. 6197--6206.

\bibitem{resnet}
K.~He, X.~Zhang, S.~Ren, and J.~Sun, ``Deep residual learning for image recognition,'' 2015.

\bibitem{mv2}
M.~Sandler, A.~Howard, M.~Zhu, A.~Zhmoginov, and L.-C. Chen, ``Mobilenetv2: Inverted residuals and linear bottlenecks,'' 2019.

\bibitem{mv3}
A.~Howard, M.~Sandler, G.~Chu, L.-C. Chen, B.~Chen, M.~Tan, W.~Wang, Y.~Zhu, R.~Pang, V.~Vasudevan, Q.~V. Le, and H.~Adam, ``Searching for mobilenetv3,'' 2019.

\bibitem{unet_pp}
Z.~Zhou, M.~M.~R. Siddiquee, N.~Tajbakhsh, and J.~Liang, ``Unet++: Redesigning skip connections to exploit multiscale features in image segmentation,'' \emph{IEEE Transactions on Medical Imaging}, 2019.

\bibitem{TD_layer}
\BIBentryALTinterwordspacing
G.~Wetzstein, D.~Lanman, M.~Hirsch, and R.~Raskar, ``Tensor displays: Compressive light field synthesis using multilayer displays with directional backlighting,'' \emph{ACM Trans. Graph.}, vol.~31, no.~4, jul 2012. [Online]. Available: \url{https://doi.org/10.1145/2185520.2185576}
\BIBentrySTDinterwordspacing

\bibitem{bridging_the_gap_bw_ViT_and_CNNs}
\BIBentryALTinterwordspacing
Z.~Lu, H.~Xie, C.~Liu, and Y.~Zhang, ``Bridging the gap between vision transformers and convolutional neural networks on small datasets,'' in \emph{Advances in Neural Information Processing Systems}, S.~Koyejo, S.~Mohamed, A.~Agarwal, D.~Belgrave, K.~Cho, and A.~Oh, Eds., vol.~35.\hskip 1em plus 0.5em minus 0.4em\relax Curran Associates, Inc., 2022, pp. 14\,663--14\,677. [Online]. Available: \url{https://proceedings.neurips.cc/paper_files/paper/2022/file/5e0b46975d1bfe6030b1687b0ada1b85-Paper-Conference.pdf}
\BIBentrySTDinterwordspacing

\bibitem{depsepconv_xception}
F.~Chollet, ``Xception: Deep learning with depthwise separable convolutions,'' in \emph{Proceedings of the IEEE Conference on Computer Vision and Pattern Recognition (CVPR)}, July 2017.

\bibitem{Sinong_lowrank}
S.~Wang, B.~Z. Li, M.~Khabsa, H.~Fang, and H.~Ma, ``Linformer: Self-attention with linear complexity,'' 2020.

\bibitem{Dong_rank_collapse}
\BIBentryALTinterwordspacing
Y.~Dong, J.-B. Cordonnier, and A.~Loukas, ``Attention is not all you need: Pure attention loses rank doubly exponentially with depth,'' \emph{ArXiv}, vol. abs/2103.03404, 2021. [Online]. Available: \url{https://api.semanticscholar.org/CorpusID:232134936}
\BIBentrySTDinterwordspacing

\bibitem{AdaBinsBhat}
S.~F. Bhat, I.~Alhashim, and P.~Wonka, ``Adabins: Depth estimation using adaptive bins,'' in \emph{Proceedings of the IEEE/CVF Conference on Computer Vision and Pattern Recognition (CVPR)}, June 2021, pp. 4009--4018.

\bibitem{invWarpOperator_NIPS}
\BIBentryALTinterwordspacing
M.~Jaderberg, K.~Simonyan, A.~Zisserman, and k.~kavukcuoglu, ``Spatial transformer networks,'' in \emph{Advances in Neural Information Processing Systems}, C.~Cortes, N.~Lawrence, D.~Lee, M.~Sugiyama, and R.~Garnett, Eds., vol.~28.\hskip 1em plus 0.5em minus 0.4em\relax Curran Associates, Inc., 2015. [Online]. Available: \url{https://proceedings.neurips.cc/paper_files/paper/2015/file/33ceb07bf4eeb3da587e268d663aba1a-Paper.pdf}
\BIBentrySTDinterwordspacing

\bibitem{pytorch}
A.~Paszke, S.~Gross, F.~Massa, A.~Lerer, J.~Bradbury, G.~Chanan, T.~Killeen, Z.~Lin, N.~Gimelshein, L.~Antiga, A.~Desmaison, A.~Kopf, E.~Yang, Z.~DeVito, M.~Raison, A.~Tejani, S.~Chilamkurthy, B.~Steiner, L.~Fang, J.~Bai, and S.~Chintala, ``Pytorch: An imperative style, high-performance deep learning library,'' in \emph{Advances in Neural Information Processing Systems 32}.\hskip 1em plus 0.5em minus 0.4em\relax Curran Associates, Inc., 2019, pp. 8024--8035.

\bibitem{adamW}
I.~Loshchilov and F.~Hutter, ``Decoupled weight decay regularization,'' 2019.

\bibitem{scheduler}
L.~N. Smith and N.~Topin, ``Super-convergence: Very fast training of neural networks using large learning rates,'' 2018.

\bibitem{deeplens2018}
W.~Lijun, S.~Xiaohui, Z.~Jianming, W.~Oliver, L.~Zhe, H.~Chih-Yao, K.~Sarah, and L.~Huchuan, ``Deeplens: Shallow depth of field from a single image,'' \emph{ACM Trans. Graph. (Proc. SIGGRAPH Asia)}, vol.~37, no.~6, pp. 6:1--6:11, 2018.

\bibitem{wang2017light}
T.-C. Wang, J.-Y. Zhu, N.~K. Kalantari, A.~A. Efros, and R.~Ramamoorthi, ``Light field video capture using a learning-based hybrid imaging system,'' \emph{ACM Transactions on Graphics (TOG)}, vol.~36, no.~4, pp. 1--13, 2017.

\bibitem{liff_stanford}
D.~G. Dansereau, B.~Girod, and G.~Wetzstein, ``Liff: Light field features in scale and depth,'' 2019.

\bibitem{ssim}
Z.~Wang, A.~Bovik, H.~Sheikh, and E.~Simoncelli, ``Image quality assessment: from error visibility to structural similarity,'' \emph{IEEE Transactions on Image Processing}, vol.~13, no.~4, pp. 600--612, 2004.

\end{thebibliography}








\end{document}